\documentclass[final,3p,times,twocolumn]{elsarticle}

\usepackage{lineno}

\journal{Journal of Archaeological Science}

\newif\ifdraft
\draftfalse

\usepackage{graphbox}

\usepackage{comment}
\usepackage{url}
\usepackage{amsthm,amsmath,amssymb} 
\usepackage{color}
\usepackage{tabularx}
\usepackage{booktabs}
\usepackage{multirow}
\usepackage[normalem]{ulem}
\usepackage{caption}
\usepackage{tikz}
\usepackage{import}
\usepackage{mathtools}

\newcommand{\diff}{\mathop{}\mathrm{d}}

\newcommand{\ceil}[1]{\left\lceil #1 \right\rceil}

\usepackage[utf8]{inputenc}
\usepackage{amsmath, amssymb, latexsym}
\DeclareMathOperator{\eLU}{eLU}

\usepackage{tikz}
\usepackage{xcolor}
\definecolor{fc}{HTML}{1E90FF}
\definecolor{sig}{HTML}{1E90FF}
\definecolor{h}{HTML}{228B22}
\definecolor{bias}{HTML}{87CEFA}
\definecolor{noise}{HTML}{8B008B}
\definecolor{conv}{HTML}{FFA500}
\definecolor{pool}{HTML}{B22222}
\definecolor{up}{HTML}{B22222}
\definecolor{view}{HTML}{FFFFFF}
\definecolor{bn}{HTML}{FFD700}
\tikzset{fc/.style={black,draw=black,fill=fc,rectangle,minimum height=1cm}}
\tikzset{h/.style={black,draw=black,fill=h,rectangle,minimum height=1cm}}
\tikzset{bias/.style={black,draw=black,fill=bias,rectangle,minimum height=1cm}}
\tikzset{noise/.style={black,draw=black,fill=noise,rectangle,minimum height=1cm}}
\tikzset{conv/.style={black,draw=black,fill=conv,rectangle,minimum height=1cm}}
\tikzset{convzero/.style={black,draw=black,fill=orange,rectangle,minimum height=1cm}}
\tikzset{sigmoid/.style={black,draw=black,fill=sig,rectangle,minimum height=1cm}}
\tikzset{drop/.style={black,draw=black,fill=pool,rectangle,minimum height=1cm}}
\tikzset{up/.style={black,draw=black,fill=up,rectangle,minimum height=1cm}}
\tikzset{view/.style={black,draw=black,fill=view,rectangle,minimum height=1cm}}
\tikzset{bn/.style={black,draw=black,fill=bn,rectangle,minimum height=1cm}}
 
\usepackage{xspace}

\usepackage[skip=5pt]{subcaption}
\newcommand{\RR}{\mathbb{R}}
\usepackage{bm}

\newcommand{\xbold}{\bm{x}}
\newcommand{\zbold}{\bm{z}}
\newcommand{\Ibold}{\mathbf{I}}

\renewcommand{\mod}[1]{\textcolor{black}{#1}}

\newcommand{\tocheck}[1]{#1}

\newcommand{\blank}{\,{\cdot}\,}
\DeclareMathOperator{\sigmoid}{sig}

\providecommand*{\eu}{\ensuremath{\mathrm{e}}} %

\begin{document}

\begin{frontmatter}
\title{Unsupervised Clustering of Roman Potsherds via Variational Autoencoders}

\author[inst1]{Simone Parisotto\corref{cor1}}
\ead{sp751@cam.ac.uk}
\author[inst2]{Ninetta Leone}
\ead{nl343@cam.ac.uk}
\author[inst1]{Carola-Bibiane Schönlieb}
\ead{cbs31@cam.ac.uk}
\author[inst2]{Alessandro Launaro}
\ead{al506@cam.ac.uk}

\cortext[cor1]{Corresponding author}

\affiliation[inst1]{organization={Department of Applied Mathematics and Theoretical Physics, University of Cambridge},
            addressline={Wilberforce Road}, 
            city={Cambridge},
            postcode={CB3 0WA}, 
            country={UK}}

\affiliation[inst2]{organization={Faculty of Classics, University of Cambridge},
            addressline={Sidgwick Avenue}, 
            city={Cambridge},
            postcode={CB3 9DA}, 
            country={UK}}

\begin{abstract}
In this paper we propose an artificial intelligence imaging solution to support archaeologists in the classification task of Roman commonware potsherds.
Usually, each potsherd is represented by its sectional profile as a two dimensional black-white image and printed in archaeological books related to specific archaeological excavations. The partiality and handcrafted variance of the fragments make their matching a challenging problem: \tocheck{we propose to pair similar profiles} via the unsupervised hierarchical clustering of non-linear features learned in the latent space of a deep convolutional Variational Autoencoder (VAE) network. Our contribution also include the creation of a ROman COmmonware POTtery (ROCOPOT) database, with \tocheck{more than} 4000 potsherds profiles extracted from 25 Roman pottery corpora, and a MATLAB GUI software for the easy inspection of shape similarities. Results are commented both from a mathematical and archaeological perspective so as to unlock new research directions in both communities.
\end{abstract}


\end{frontmatter}

\section{Introduction}
Our ability to \emph{interpret} an archaeological site rests on our capacity to recognise the material culture found in it, ranging from fixed structures to movable objects. Archaeological finds from any site do in fact contribute to defining its chronology, function and place within a broader network of relationships with other sites. Thanks to their relative resilience against decay combined with their specific underlying patterns of production, distribution and consumption, ceramic \emph{vessels} (pottery) – whether fragmented (i.e.\ potsherds) or intact – represent some of the most common finds recovered during archaeological fieldwork. Individual pots and potsherds are usually recorded as 2D profiles and, in consideration of their morphological features (e.g.\ shape of the rim or the base), gathered in systematic catalogues (\emph{corpora}), where patterns of similarity are used to establish relationships, in terms of function, chronology or both \cite{orton_hughes_2013}.

Within Roman archaeology specifically (albeit not exclusively), fundamental significance has been ascribed to those ceramic classes more closely associated with long-distance trade and contact, namely trade containers (\emph{amphorae}) and high-quality tableware pots (\emph{finewares}). However, these were middle-range commodities that did not reach all levels of the society in the same way, and, furthermore, their supply varied enormously over space and time. Indeed, some sites might have had limited or no access to this range of objects, and such notable absence in the archaeological record might indicate their abandonment at a time when they were in full occupation (i.e.\ reduced archaeological visibility).
In contrast, ordinary table-- and kitchen--ware (\emph{commonwares}, also referred to as \emph{coarsewares}) were considerably cheaper and mainly supplied within a local/regional network of distribution. As a result, they almost invariably constitute the bulk of pottery finds at almost every Roman site. Even though one would expect them to provide a most effective baseline for the dating (and more general interpretation) of Roman sites anywhere, their huge range of forms, poorly defined chronologies and scattered provenance have made their study so challenging (and so little promising) that many have favoured the analysis of the far more standardised and easily recognisable finewares and amphorae.

\begin{figure}[htb]
    \centering
    \includegraphics[width=0.47\textwidth,trim=5.25cm 5cm 4.75cm 6.25cm,clip=true]{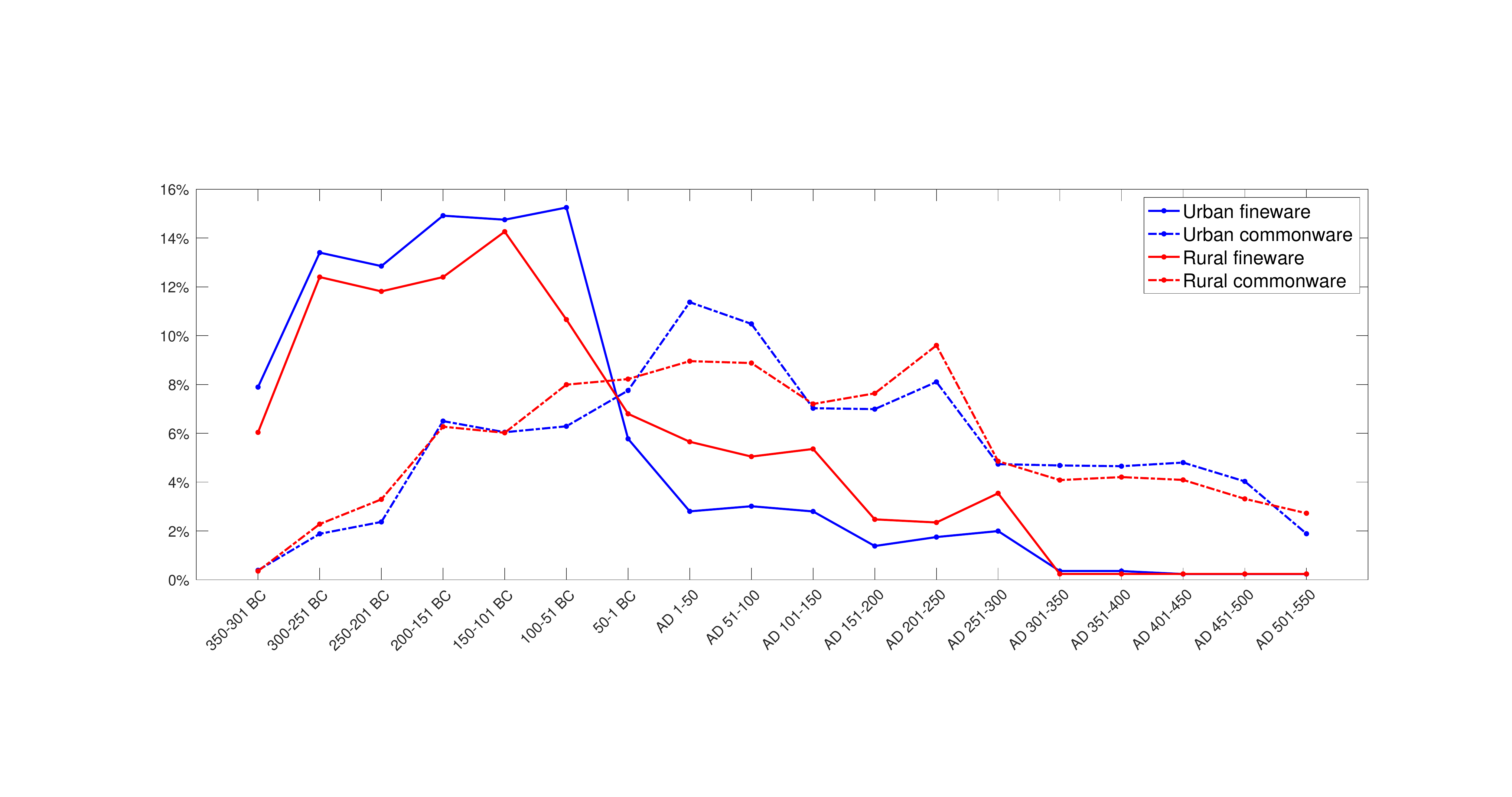}
    \caption{Percentage of fineware (straight) and commonware (dashed) potsherds: urban (blue) vs.\ rural (red) environment.}
    \label{fig: fineware vs commonware}
\end{figure}

Nevertheless, when a special effort is made to include a comprehensive study of commonware sherds, resulting interpretations can dramatically change. This was neatly shown by a recent analysis of the chronological distribution of potsherds recovered from the Roman town of Interamna Lirenas and from across its surrounding countryside (Central Italy) \cite{LauLeo2018}. Whereas traditional reliance on finewares and amphorae had outlined a trajectory of precocious demographic decline (already in progress by the late 1st century BC), the widespread presence of commonwares shows a considerably more gradual process of growth, in fact peaking in the course of the 1st century AD, with little or no sign of decline until about two centuries later, see Figure \ref{fig: fineware vs commonware}. 
There can be no doubt that the study of commonwares should be a priority within (Roman) material culture studies. However, this does not make the obstacles which archaeologists have to face any less real. What is indeed needed is to improve and expand our classification of this vast and varied body of evidence in a way which is considerably more effective and less time-consuming than it currently is.

\paragraph{Contributions}

\begin{figure*}[htb]
    \centering
    \includegraphics[width=1\textwidth]{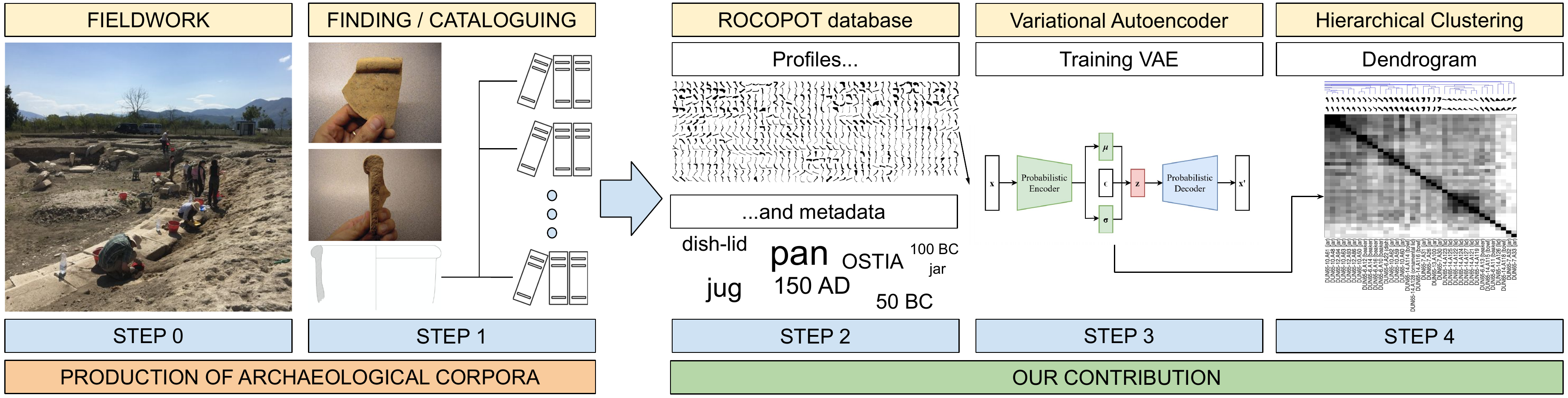}
    \caption{Proposed workflow for clustering Roman potsherds. 
    }
    \label{fig: workflow}
\end{figure*}

The contribution of this paper is two-fold. 
Firstly, we create the \emph{ROman COmmonware POTtery} (ROCOPOT) database containing two-dimensional black-white imaging profiles of commonware potsherds attested in Central Tyrrhenian Italy, with metadata extrapolated from different corpora and excavation sites. 
Secondly, we propose an unsupervised workflow for matching and hierarchically clustering the potsherds profiles by comparing their latent representation learned in a deep convolutional Variational Autoencoder (VAE) network, and supported by a MATLAB GUI software for the easy inspection of the results on the field.

\paragraph{Scope}
\tocheck{Our research aims to support archaeologists in the classification task of Roman commonware potsherds by unveiling new similarity patterns between individual profiles as featured in relevant corpora, especially those for which no match has so far been proposed. The resulting clustering patterns \tocheck{will} highlight new morphological relationships and serve as an additional tool for archaeologists in order to improve their understanding of chronology, distribution, function and development of Roman commonwares over time. 
Whereas other similar projects have attempted to develop automated procedures for the correct identification of potsherds (i.e.\ establishing a relationship between what is newly found and the published 2D profiles), we aim to improve the internal consistency of the database itself and facilitate the archaeological classification workflow. 
On top of that, our workflow is fully scalable, meaning that once new profiles are available as black-white images, the learned representation features will unlock additional matches.}
The complete pipeline, from archaeological fieldwork to our contributions, is reported in Figure \ref{fig: workflow}.


\paragraph{Related works}
Shape recognition, classification and matching of complete or partial data, are well-established problems in the computer vision and mathematical communities but for settled databases of thousands of images or shapes, e.g.\ \cite{Bronstein2009,MNIST,Shilane:2004:TPS}.
In recent years, ``Cultural Heritage Imaging Science'' became an increasing popular research field bringing together experts from museum institutions, history of art, physics, chemistry, computer vision and mathematics departments for tackling cross-discipline challenging applications;
however, few dedicated databases emerged in the literature, mainly related to collection of paintings and associated tasks, e.g.\ for object or people detection \cite{CroZis2015,GinHasBroMal2015,Gonthier18}, style recognition \cite{BarLevWol2015} and many more, see \cite{FioKhoPonTraDBue2020}. 

Relevant works for matching 2D shapes are based on similarities of boundaries \cite{CuiFemHuWonRaz2009,SunSup2005} or skeletons \cite{ShenWangBaiWangJan2013}, homeomorphic transformations based on size functions \cite{dAmFroLan2006}, geodesic calculus for image morphing \cite{RumWir2013}, axiomatic criteria for deformable shapes \cite{Bronstein2007} or comparison of invariant image moments \cite{KimKim2000,ManCreByuYezSoa2006,Hu1962,ZunZun2014}. 
In particular, invariant shape descriptors are detailed in \cite{Cao2008} for clustering together similar shapes while the meaningfulness of the obtained clusters is still an open question. 
For 3D surfaces (with triangular meshes) or volumes (with voxels), remarkable results are obtained via scale invariant descriptors based on the eigen-decomposition of the Laplace--Beltrami operator, e.g.\ the \emph{Heat Kernel Signature} and its variants \cite{LitRodBroBro2017,RavBroBroKim2010,SunOvsGui2009}.

For the targeted application of this paper, even fewer databases are publicly available \cite{CeramAlex,Tyers:1996,university_of_southampton_roman_2014}, with focus on their automatic digitisation, shape extraction and visual presentation \cite{ArchAIDE} or the clustering of a-priori manually selected geometric shape features into similar classes, e.g.\ by comparing curve skeletons \cite{PiccoliChiara2015Ttac}, shape boundaries \cite{Smith2012}, shape descriptors \cite{SeiWieZepPinBre2015} or employing Generalised Hough Transform distance measures (in the case of petroglyphs, a dataset that shares similarity with our images) \cite{Zhu2010} and other topological features (e.g.\ pixels height and width, area, circularity, rectangularity, diameters or steepness indexes) \cite{Christmas2018,HorrLindBrun2014} or comparison with template primitives \cite{Kampel01classificationof}.
In \cite{BanDelEvaGatItkZal2017} fragments and nodal points are also extracted from complete 3D models in view of a \emph{supervised} deep learning approach starting from complete profiles.
Recently, it is worth mentioning the deep-learning approach based on convolutional neural networks (CNNs) proposed in \cite{PawDow2021}, which targeted the imaging classification of decorative styles in Tusayan White Ware (Northeast Arizona).

In contrast, our deep-learning approach is not biased by the selection of a-priori shape features, allowing for their \emph{unsupervised} extraction by means of a deep convolutional Variational Autoencoder (VAE) network. 
\mod{Here, we force the VAE latent space to follow a prior probability distribution close to a standard Gaussian. Such regularisation has many advantages for unsupervised tasks mainly related to the easy reparametrisation of the variational lower bound and backpropagation through the stochastic layers \cite{KinWel2014}. 
} Our approach is motivated by the high availability of fragments and the generative process of VAE, allowing us to extend the learned model to unseen data \mod{for extracting} their shape features. 

In pattern recognition, hierarchical clustering is a non-parametric yet versatile unsupervised approach for unveiling inherent structures in data, ordered in a tree called \emph{dendrogram} \cite{DudHarSto2000}. The method is based on the recursive partitioning of the \mod{VAE features} into clusters of (increasing or decreasing) cardinality, based on the minimisation of a certain cost function that promotes the separability of the data features \cite{CohKanMalMat2019}.
\mod{Hierarchical} clustering methods are easy to implement \mod{and} they are often tuned to the application at hand, with an external evaluation of the results by the experts in the applied field \cite{VekslerCourse2004}.

Specifically to Roman pottery profiles, our approach for an \emph{unsupervised hierarchical clustering} is in line with the promising works in \cite{HorrLindBrun2014,KarSmi2011,Smith2012,Zhu2010,NAVARRO2021}, with the additional automatisation of the cluster merging rule based on the best cophenetic coefficient score \cite{SokRoh1962}.
Finally, we leave the check of the deeper clusters in the produced dendrograms to specialist archaeologists, who can effectively assess the performances of the proposed workflow.

\paragraph{Organisation of the paper} 
The paper is organised as follows:
in Section \ref{sec: database} we introduce the \emph{ROman COmmonware POTtery} (ROCOPOT) database; 
in Section \ref{sec: problem} we detail about the deep variational autoencoder (VAE) network for extracting the potsherds features and the hierarchical clustering algorithm (with appendix on computations in Section \ref{sec: appendix});
in Section \ref{sec: results} we discuss the results obtained by our workflow.

\section{The ROCOPOT Database} \label{sec: database}
The \emph{ROman COmmoware POTtery} (ROCOPOT) database, downloadable from \cite{ROCOPOTdatabase},
comprises of more than 4000 black-white images, consisting of two-dimensional representation of the section (profile) of Roman commonware vessels. These profiles are featured in a series of archaeological \emph{corpora} \cite{BRAG96,CICI96,CIPR96,CM91,CT84,DECA94,DIGIO96,DUN64,DUN65,DYS76,FEDE96,FULF84,FULF94,GASP96,LUNI2,OLCE93,OSTIA1,OSTIA2,OSTIA3,OSTIA4,PAP85,POHL70,ROB97,SCAHO96,STAN01} which provide a representative sample of a wide array of vessel-forms attested across the Tyrrhenian side of the Italian peninsula in the Roman period (from Liguria to Campania).
In this work we present the version 1.0 of our database (future extensions are planned).
All these images are scanned at 300dpi from the archaeological corpora: these are saved in the lossless \texttt{.png} format with an identification string filename of the form \texttt{IDCAT-PAGNUM.FIGID.png}, where \texttt{IDCAT} identifies the catalogue while \texttt{PAGENUM} and \texttt{FIGID} \tocheck{are the page and the figure identification numbers of the printed profile in the corpora, respectively.}
This labelling convention is suitable for a fast inspection of the results described in Section \ref{sec: results}: by assuming similar shapes are presented closely in each catalogue (as is normally the case), we can quickly look at filenames to evaluate potential matches.

\begin{figure}[htb]
    \centering
    \includegraphics[width=0.41\textwidth]{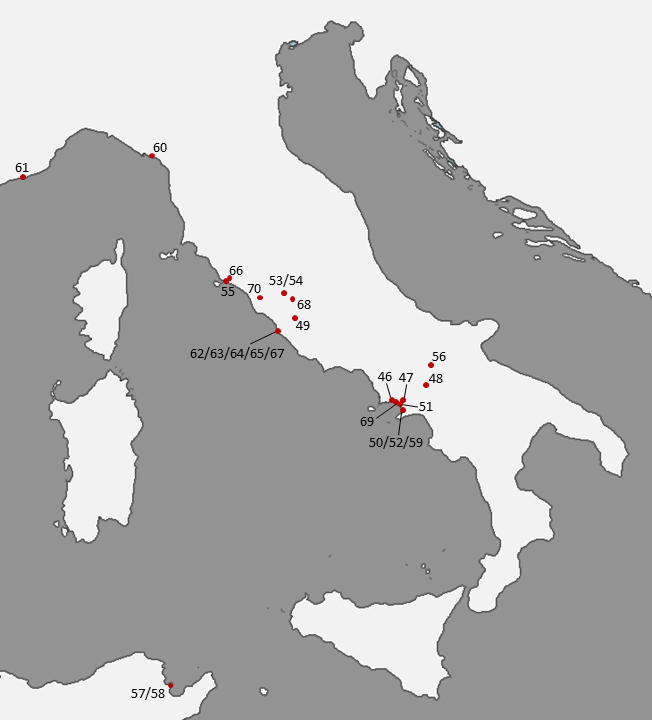}
    \caption{Location of the sites mentioned in Table \ref{tab: database 1.0}.}
    \label{fig:map}
\end{figure}

Since the original profiles can be composed of multiple (\tocheck{rarely} intact) parts, we refined the database identifying a total of 407 bases (B), 450 handles (H), 3678 rims\footnote{the rim usually refers to a rounded moulding on the lip of a jar, bowl, or dish, both to add strength and assist in handling \cite{OxfordConcise}.} (R) and 451 rims with handles (RH), as well as the original shapes (O). 
The archaeologists identified such parts out of each original profile depending on its grade of completeness. Potential outliers like one of a kind profiles were removed.
Details about the number of profiles in our database are summarised in Table \ref{tab: database 1.0} (see also the map in Figure \ref{fig:map}), where for each catalogue we highlight the bibliographic reference, the publication year, the chronology and location of the archaeological site. 
For the rest of this work we focused only on the rims as they are assumed to carry out the most significant geometric information of the handcrafted material.

\begin{figure}[!htb]
\centering
\begin{subfigure}[t]{0.47\textwidth}\centering
\includegraphics[height=1.5cm,trim=1.5cm 0cm 1.5cm 0cm,clip=true]{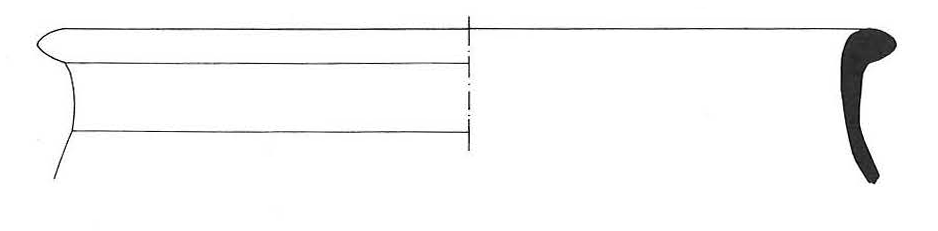}
\hfill
\includegraphics[height=1.5cm]{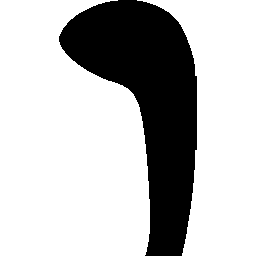}
\captionsetup{justification=centering}
\caption{DYS76-3.CF39}
\label{fig: rim separation}
\end{subfigure}
\begin{subfigure}[t]{0.47\textwidth}\centering
\includegraphics[height=2.0cm,align=m]{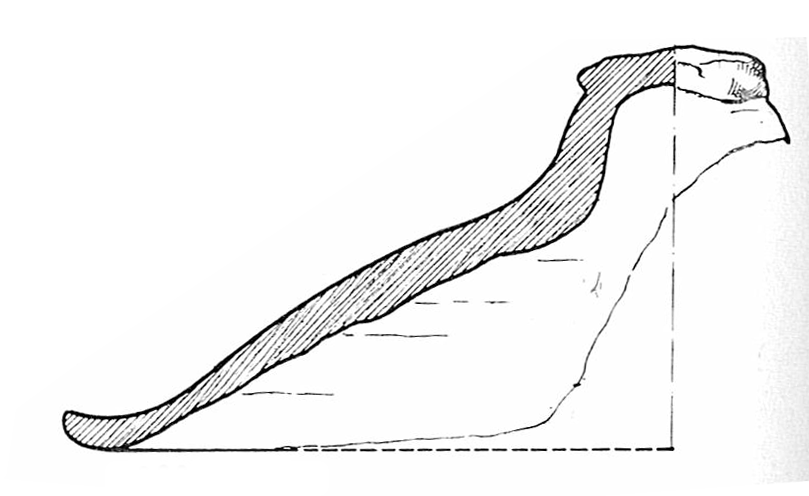}
\hfill
\includegraphics[height=1.5cm,align=m]{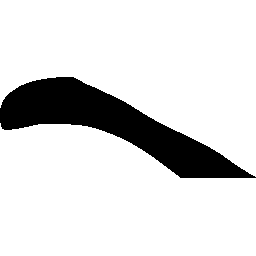}
\captionsetup{justification=centering}
\caption{POHL70-55.38}
\label{fig: rim filling 1}
\end{subfigure}
\begin{subfigure}[t]{0.47\textwidth}\centering
\includegraphics[height=2cm,trim=1em 3.1em 1em 3.5em,clip=true,align=m]{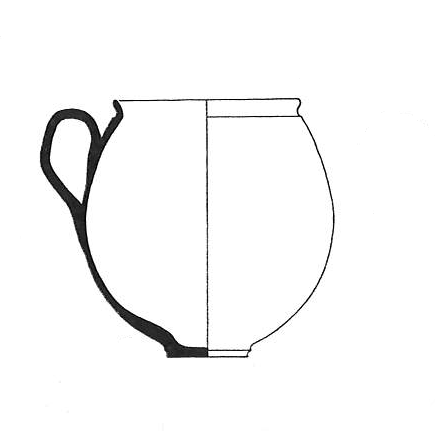}
\hfill
\includegraphics[height=1.5cm,align=m]{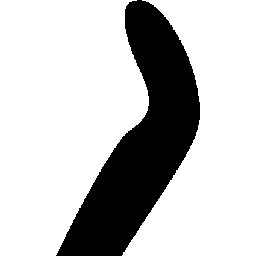}
\captionsetup{justification=centering}
\caption{DUN64-7.3}
\label{fig: rim polishing 1}
\end{subfigure}
\begin{subfigure}[t]{0.47\textwidth}\centering
\includegraphics[height=2cm,align=m]{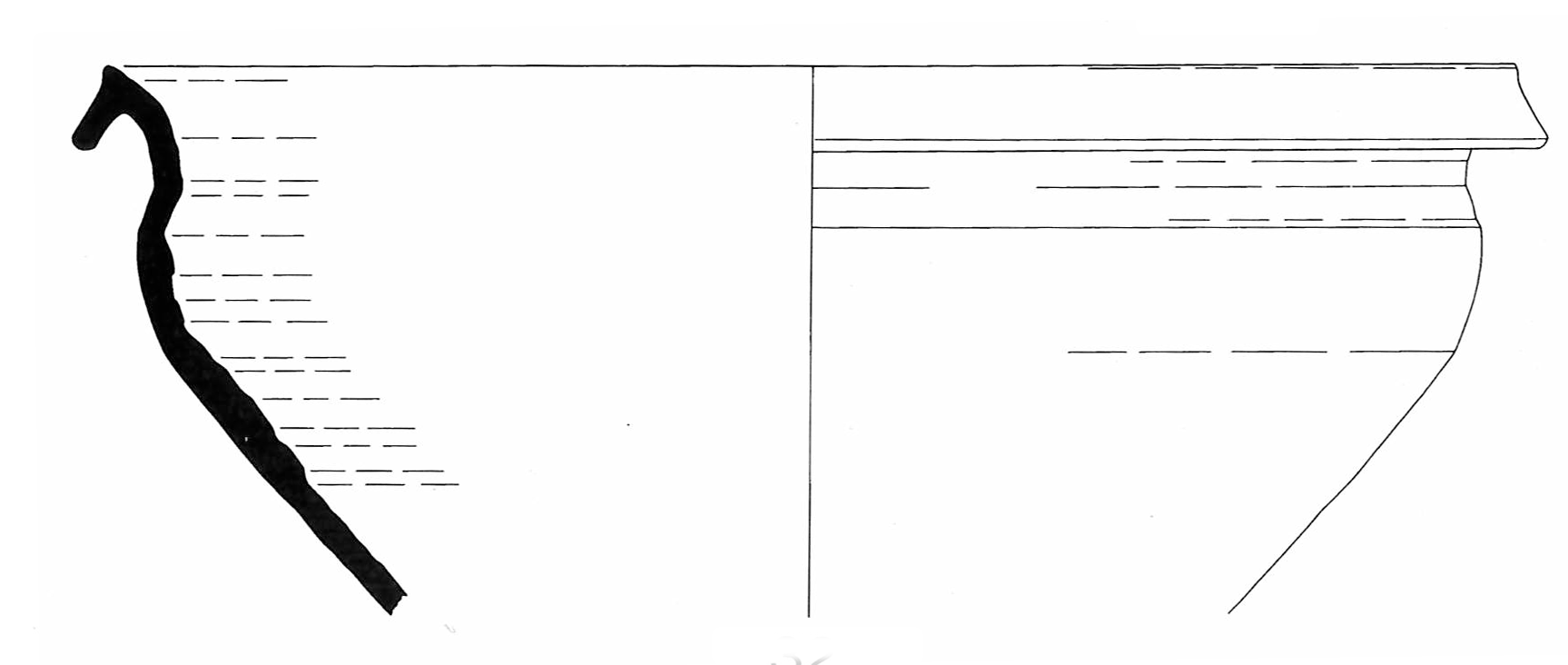}
\hfill
\includegraphics[height=1.5cm,align=m]{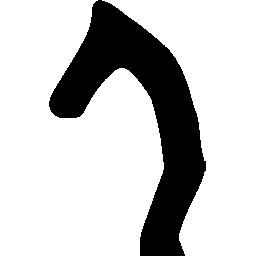}
\captionsetup{justification=centering}
\caption{ROB97-220.32}
\label{fig: rim polishing 2}
\end{subfigure}
\caption{Preprocessed shapes (right) from originals (left).}
\label{fig: preprocessing}
\end{figure}

\begin{figure*}[!htbp]\centering
\centering
\subcaptionbox{BRAG96}{
\includegraphics[height=3.8cm]{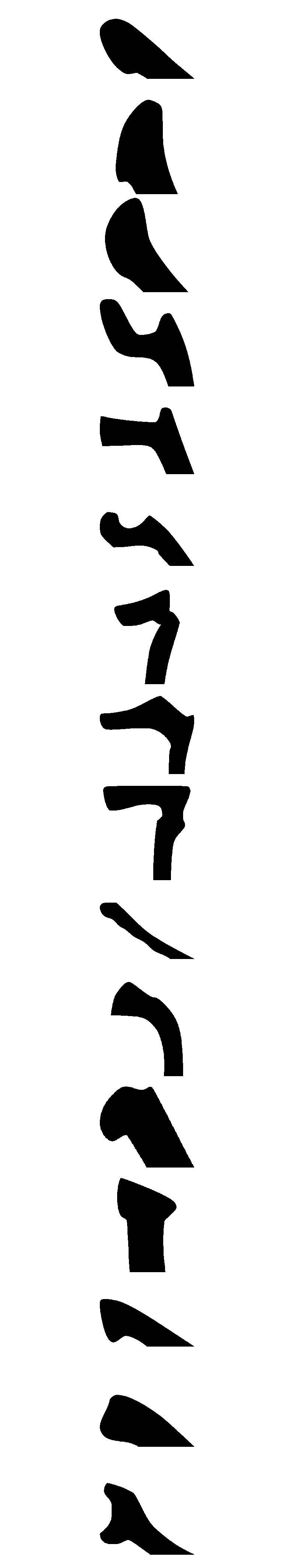}
}\quad
\subcaptionbox{CICI96}{
\includegraphics[height=3.8cm]{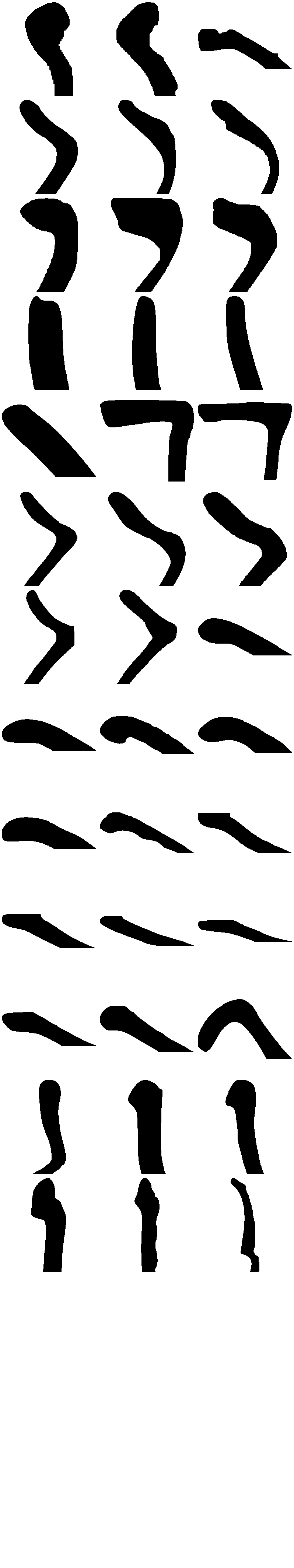}%
}\quad
\subcaptionbox{CIPR96}{
\includegraphics[height=3.8cm]{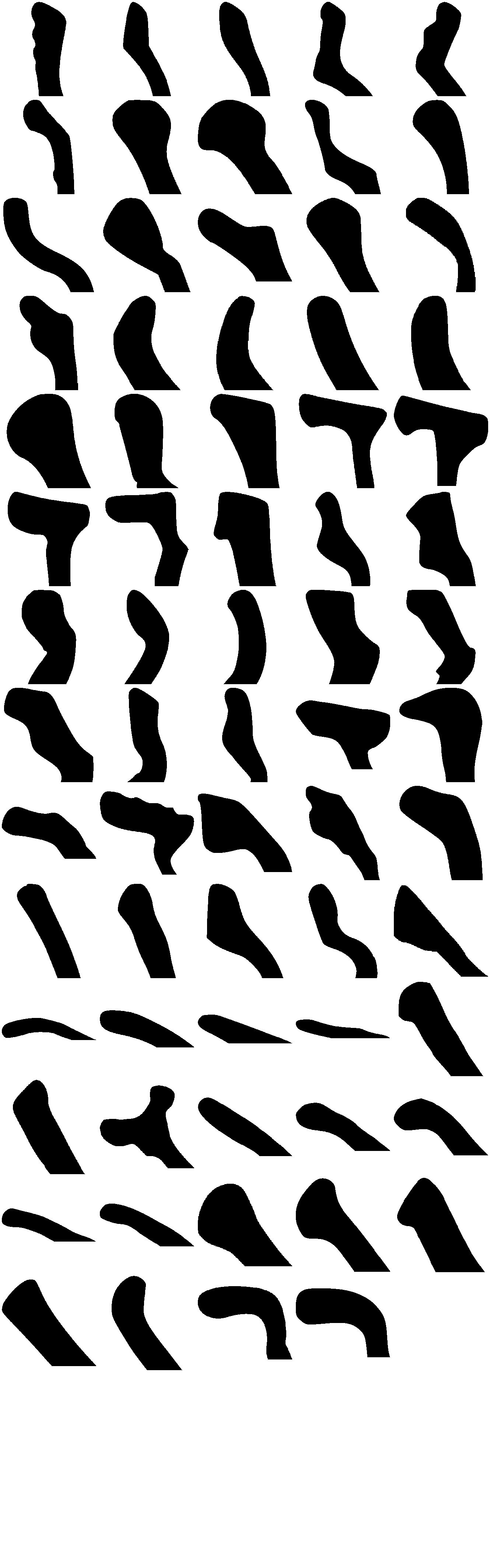}%
}\quad
\subcaptionbox{CM91}{
\includegraphics[height=3.8cm]{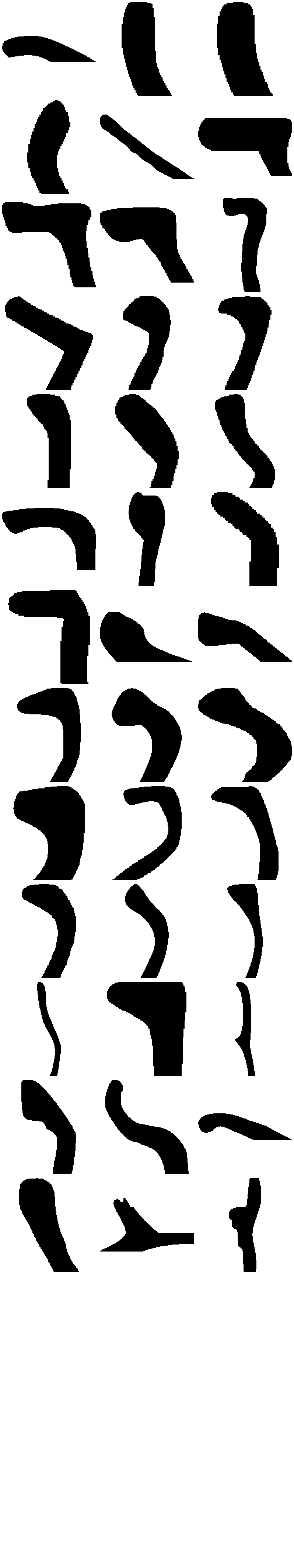}%
}\quad
\subcaptionbox{CT84}{
\includegraphics[height=3.8cm]{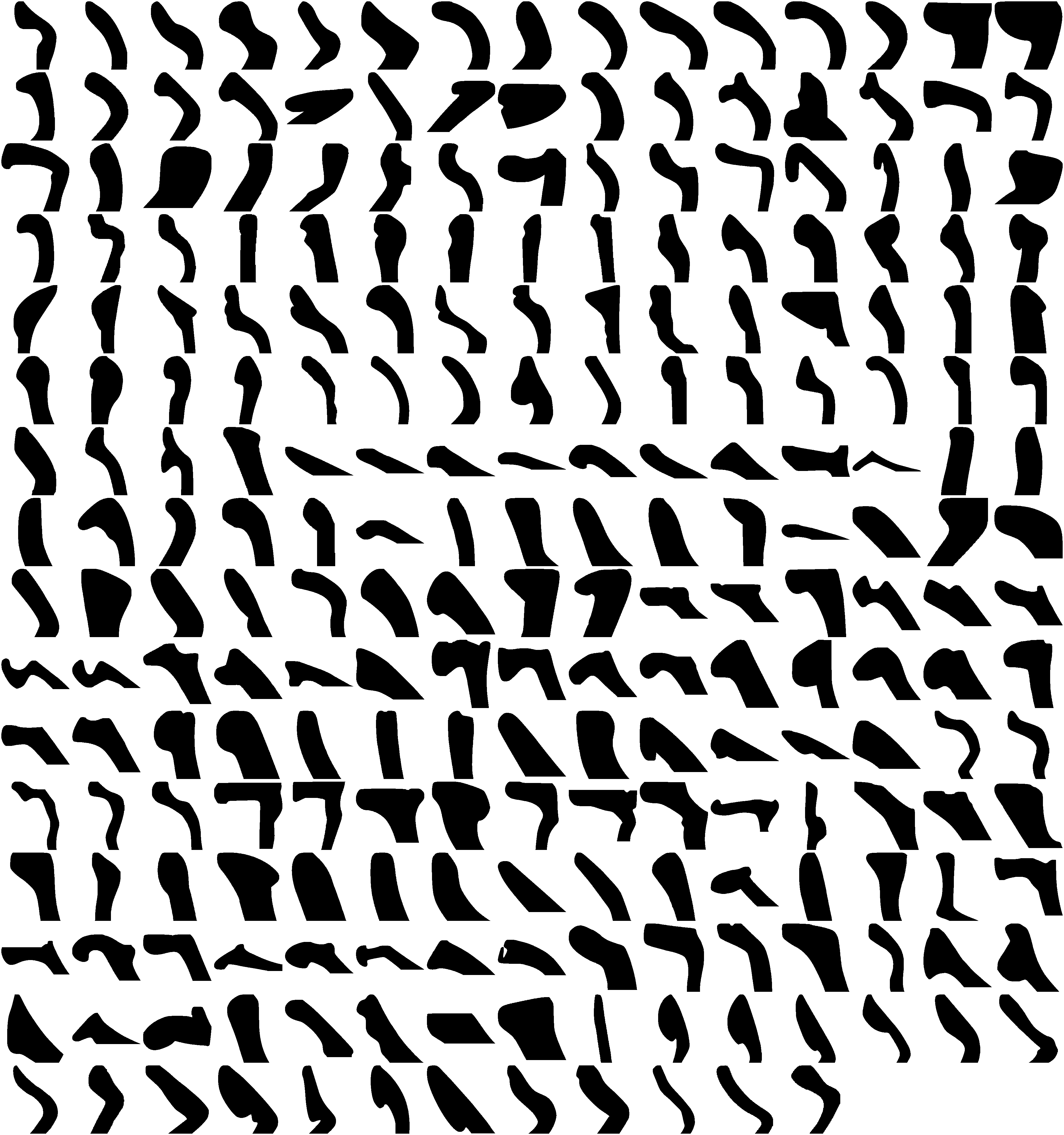}%
}\quad
\subcaptionbox{DECA94}{
\includegraphics[height=3.8cm]{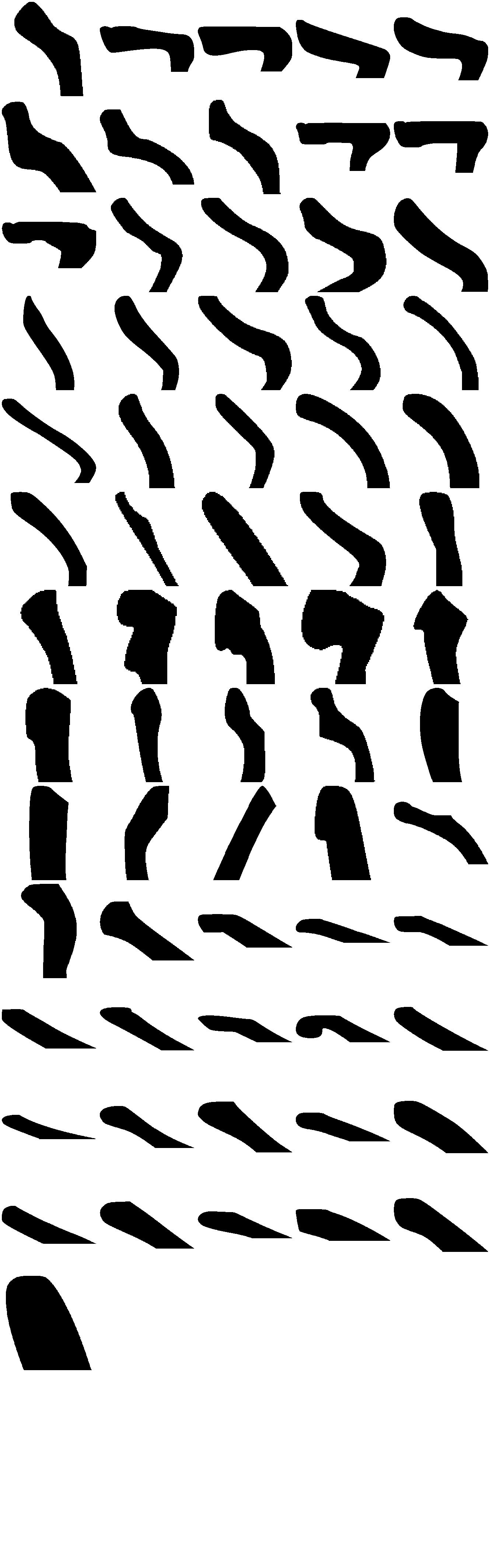}%
}\quad
\subcaptionbox{DIGIO96}{
\includegraphics[height=3.8cm]{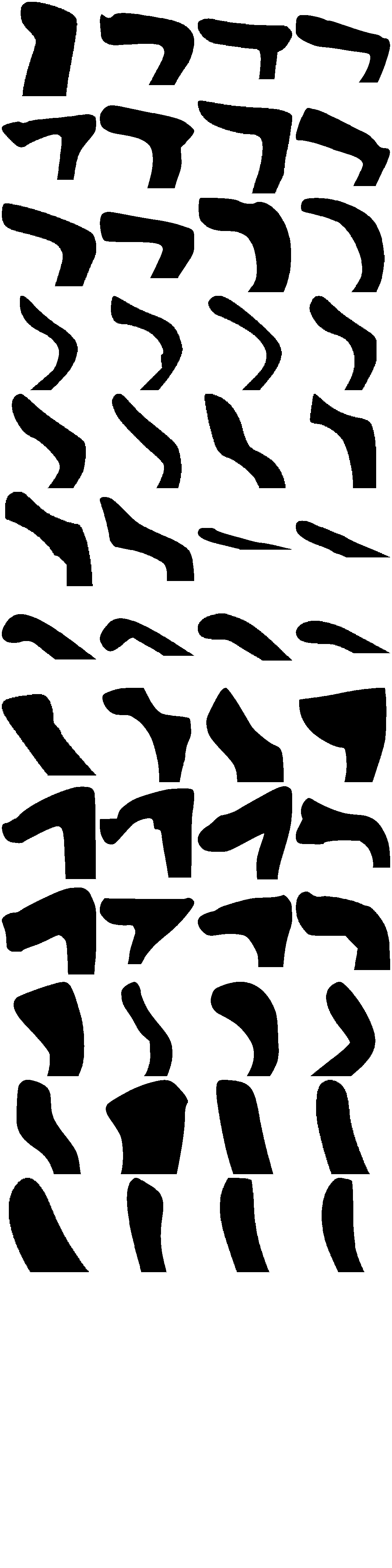}
}\quad
\subcaptionbox{DYS76}{
\includegraphics[height=3.8cm]{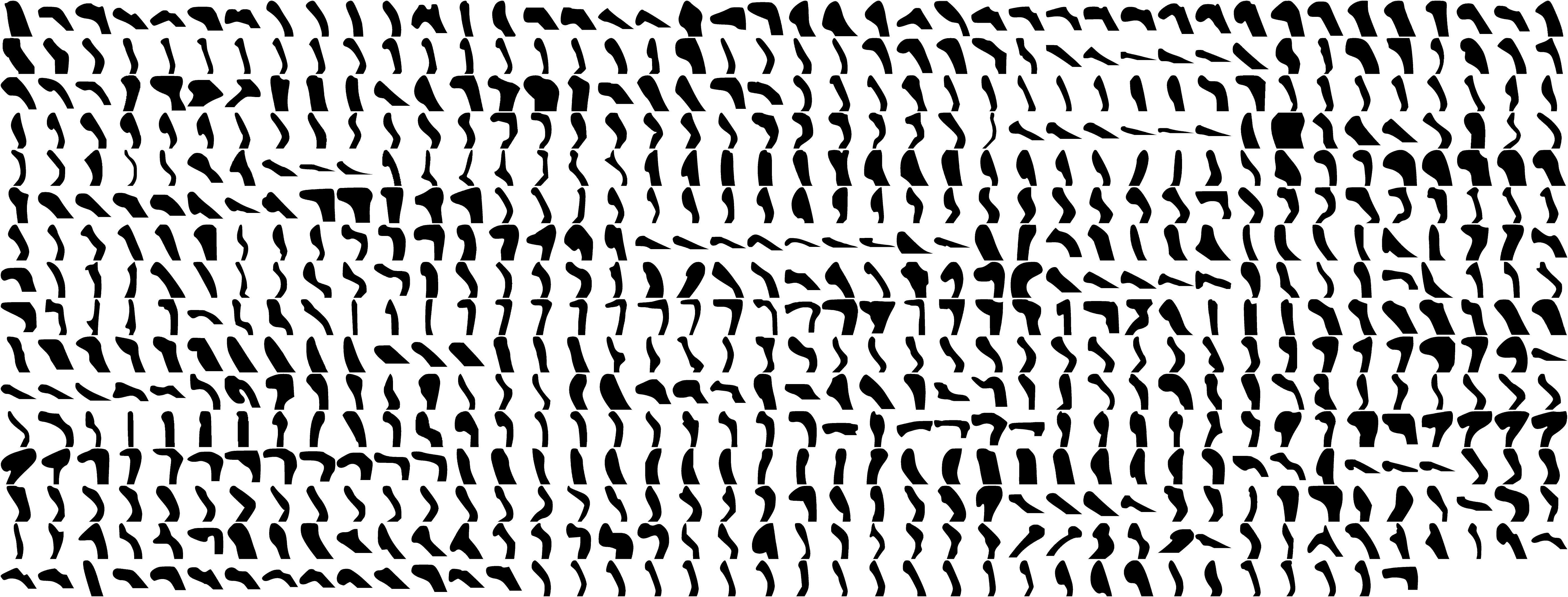}
}\quad
\subcaptionbox{DUN64}{
\includegraphics[height=3.8cm]{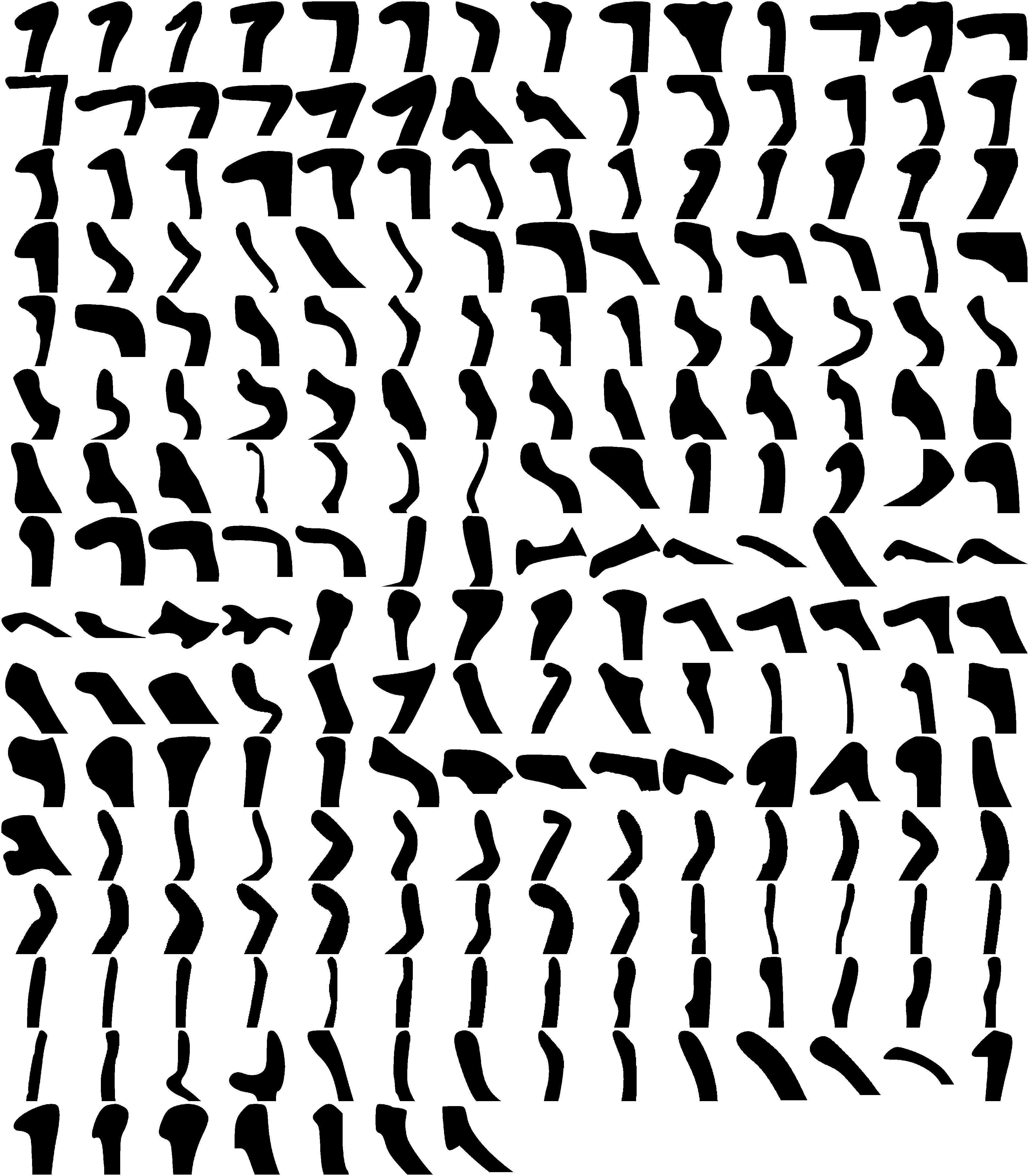}
}\quad
\subcaptionbox{DUN65}{
\includegraphics[height=3.8cm]{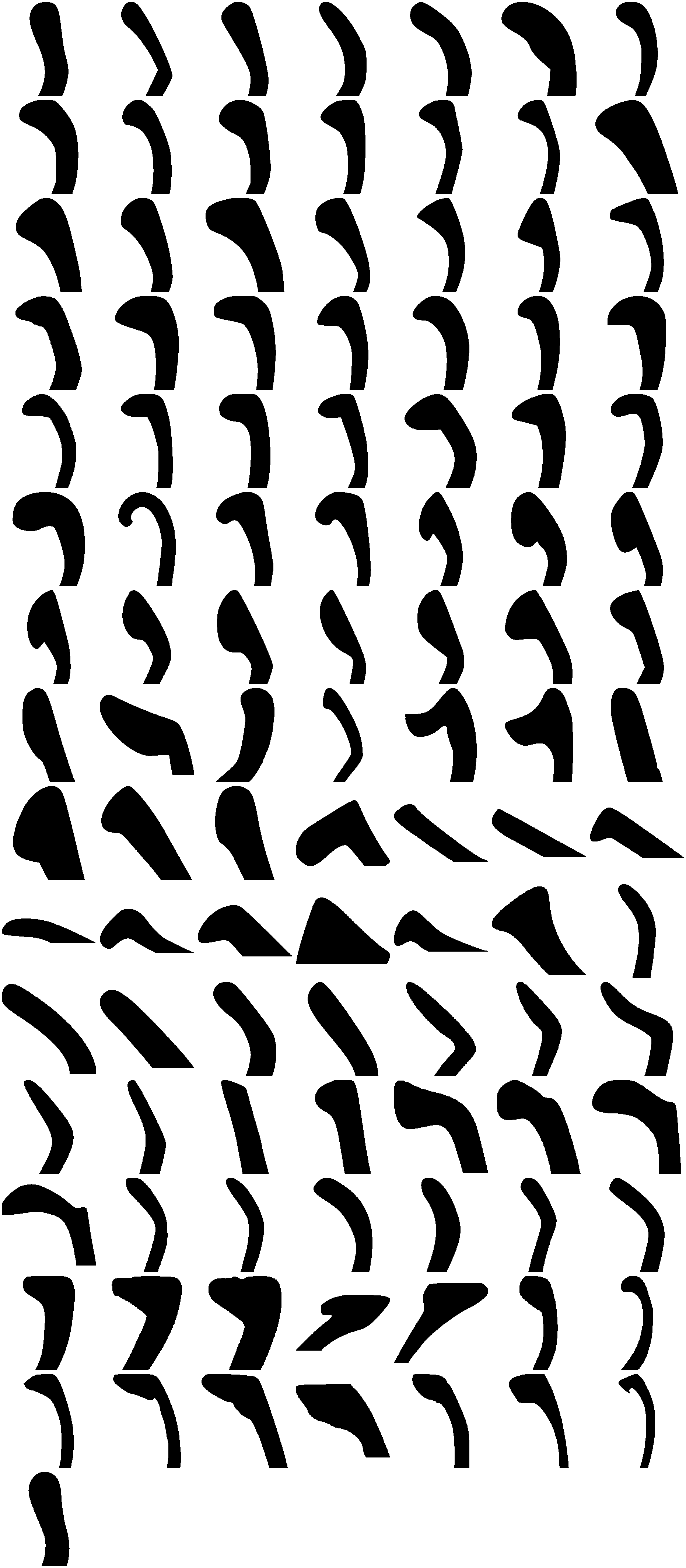}
}\quad
\subcaptionbox{FEDE96}{
\includegraphics[height=3.8cm]{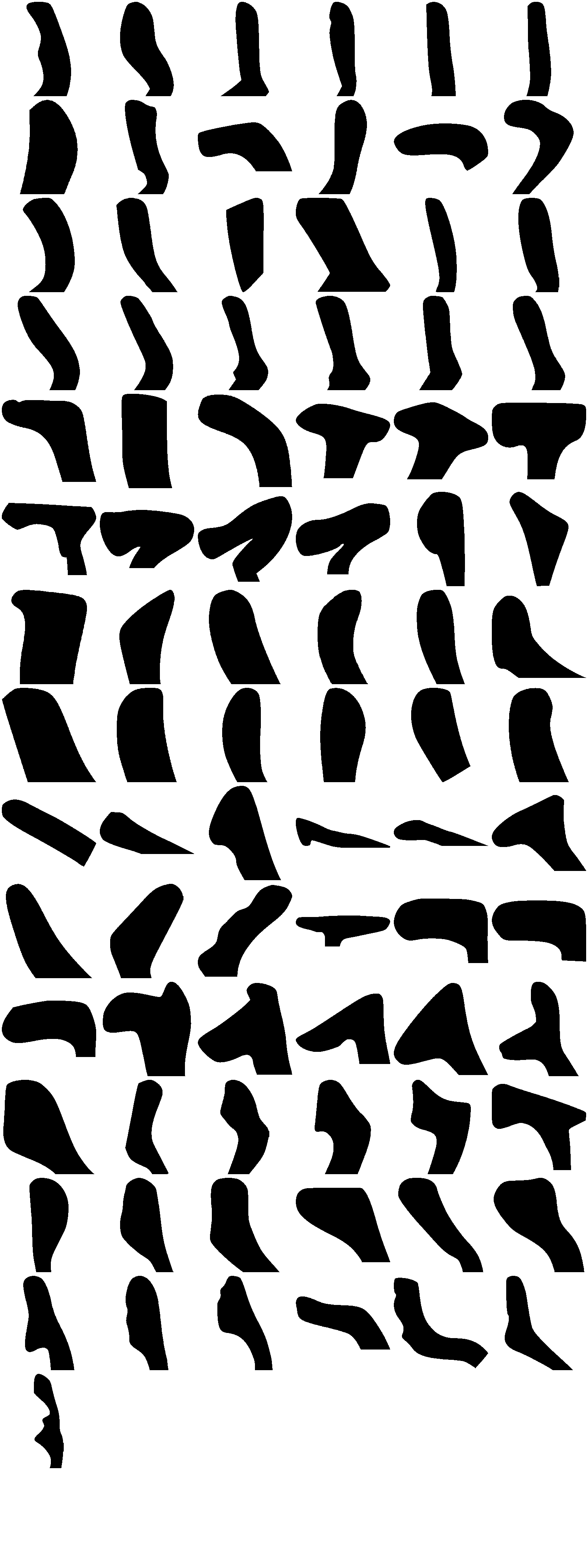}
}\quad
\subcaptionbox{FULF84}{
\includegraphics[height=3.8cm]{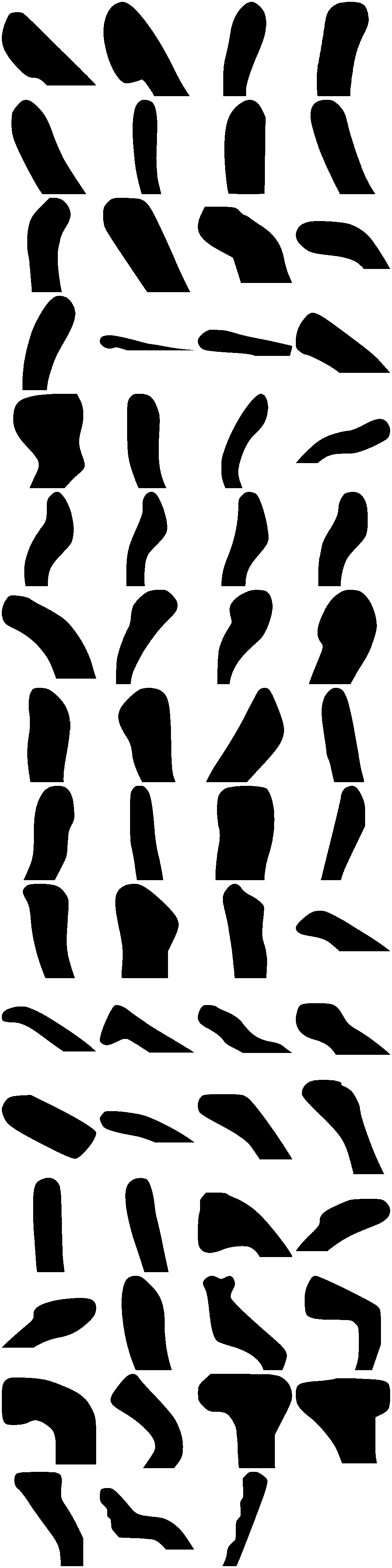}
}\quad
\subcaptionbox{FULF94}{
\includegraphics[height=3.8cm]{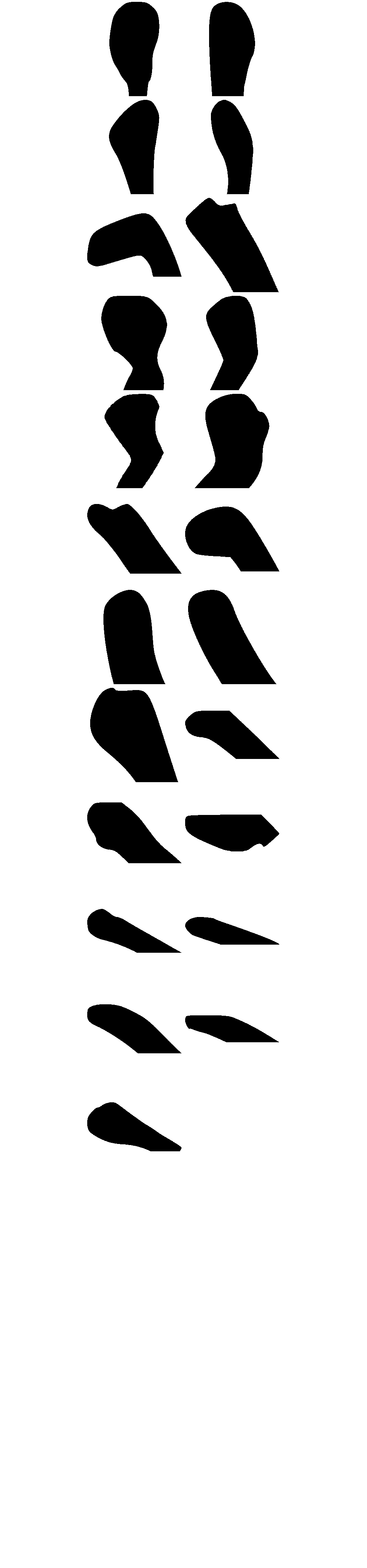}
}\quad
\subcaptionbox{GASP96}{
\includegraphics[height=3.8cm]{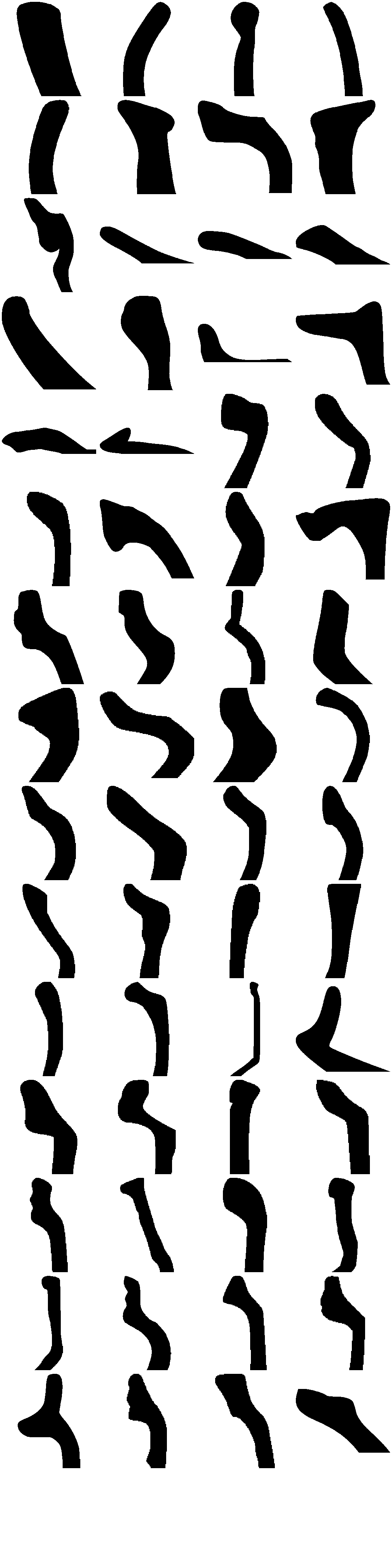}
}\quad
\subcaptionbox{LUNI2}{
\includegraphics[height=3.8cm]{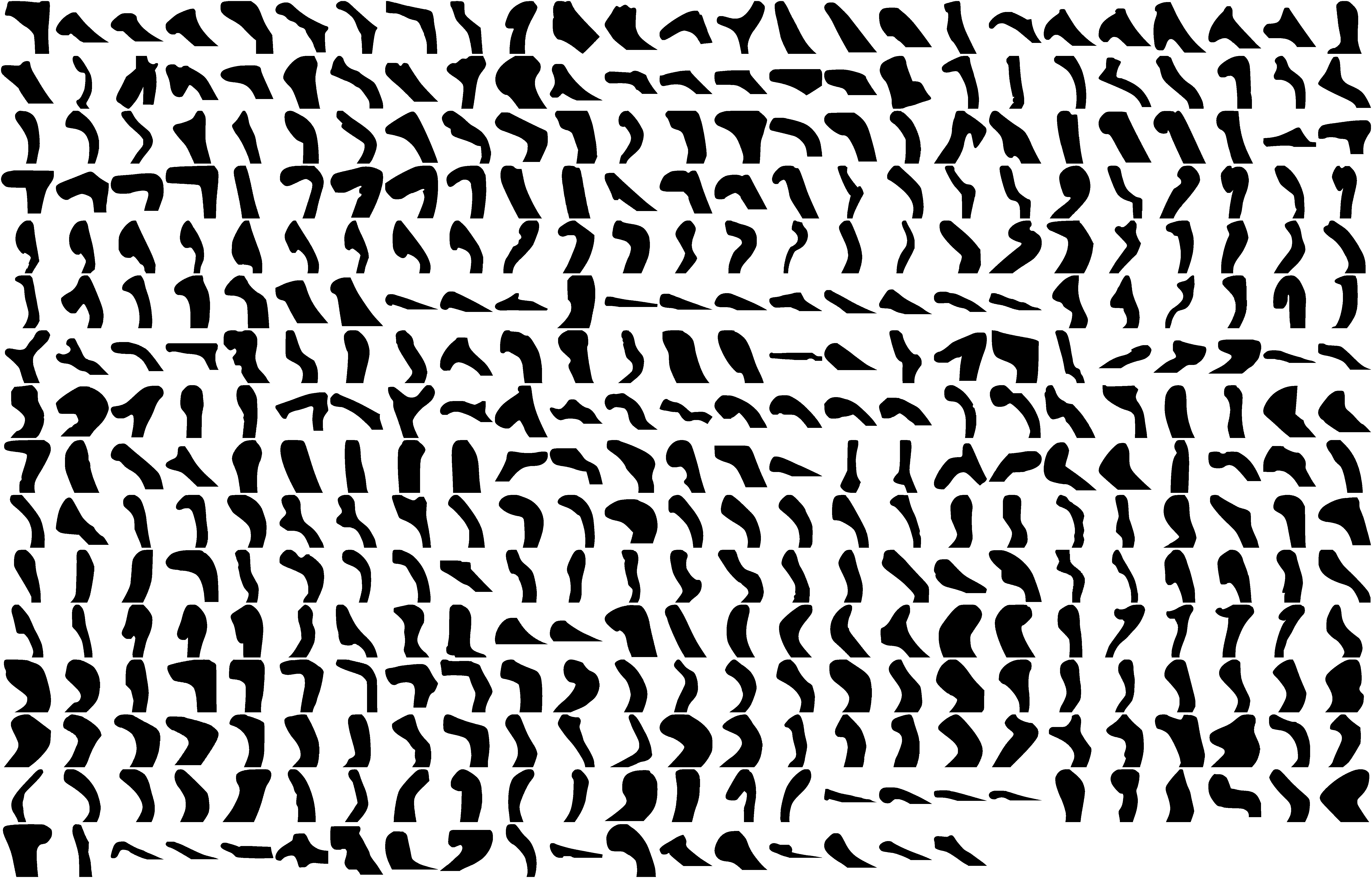}
}\quad
\subcaptionbox{OLCE93}{
\includegraphics[height=3.8cm]{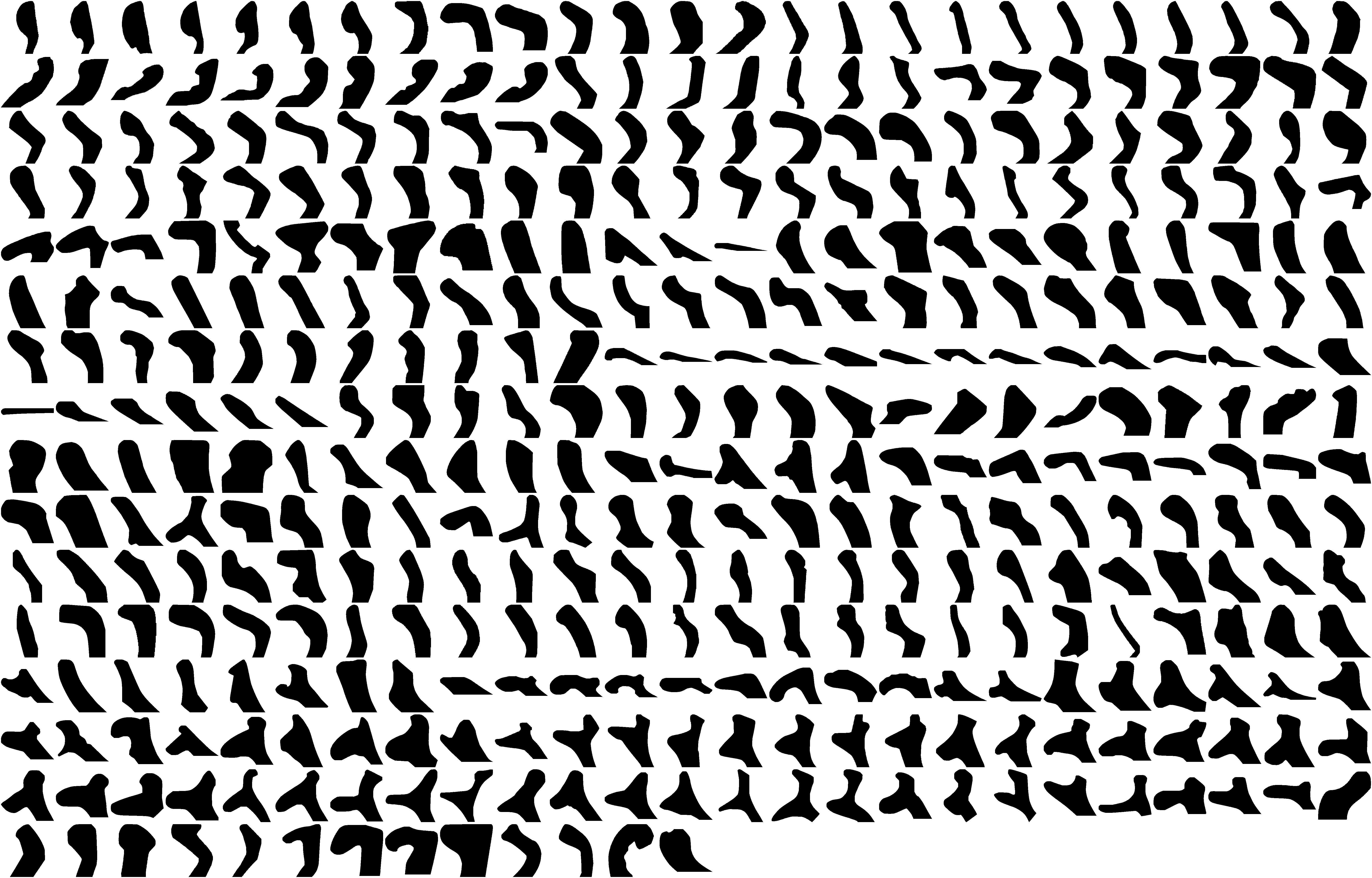}
}\quad
\subcaptionbox{OSTIA1}{
\includegraphics[height=3.8cm]{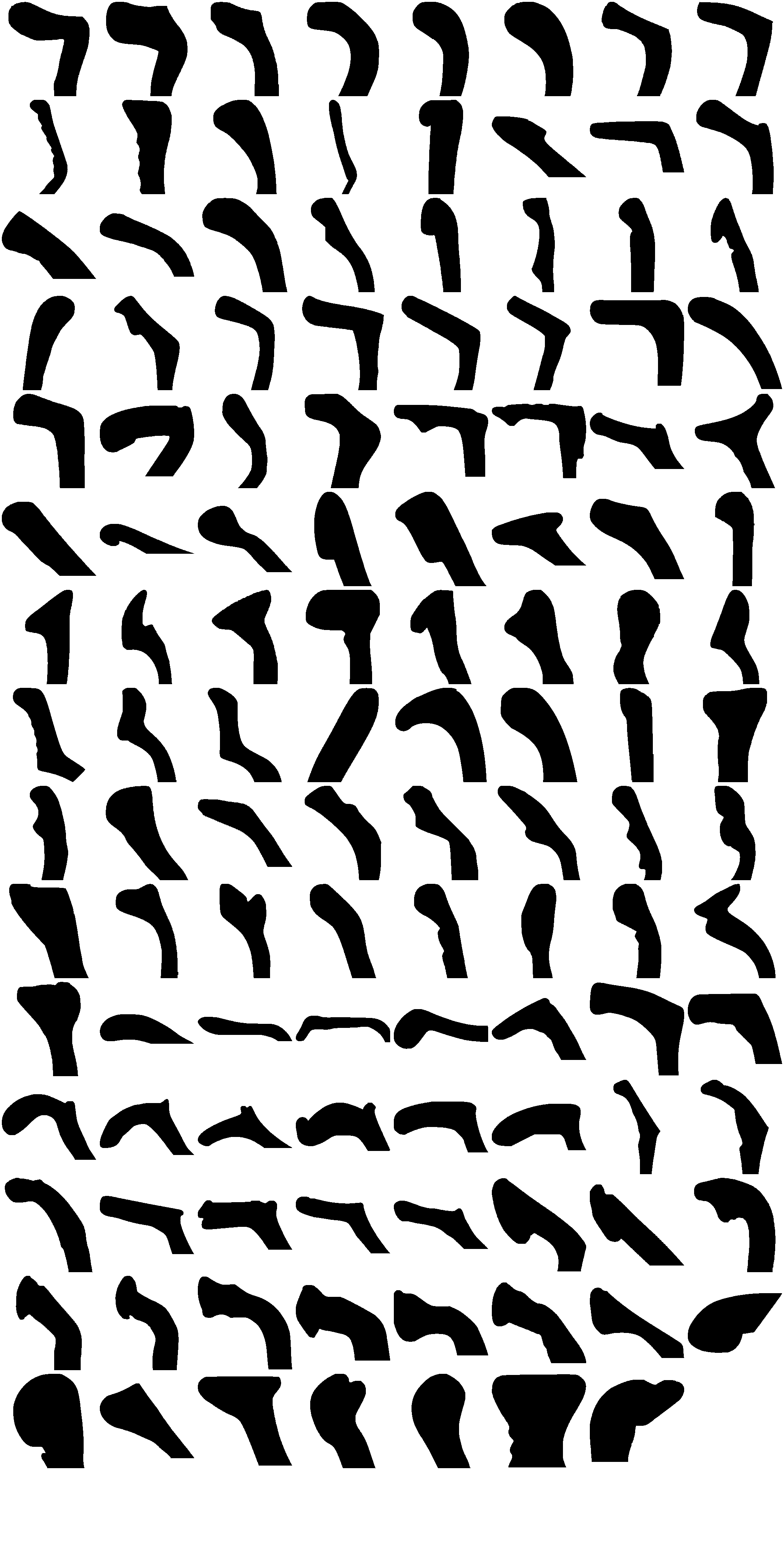}
}\quad
\subcaptionbox{OSTIA2}{
\includegraphics[height=3.8cm]{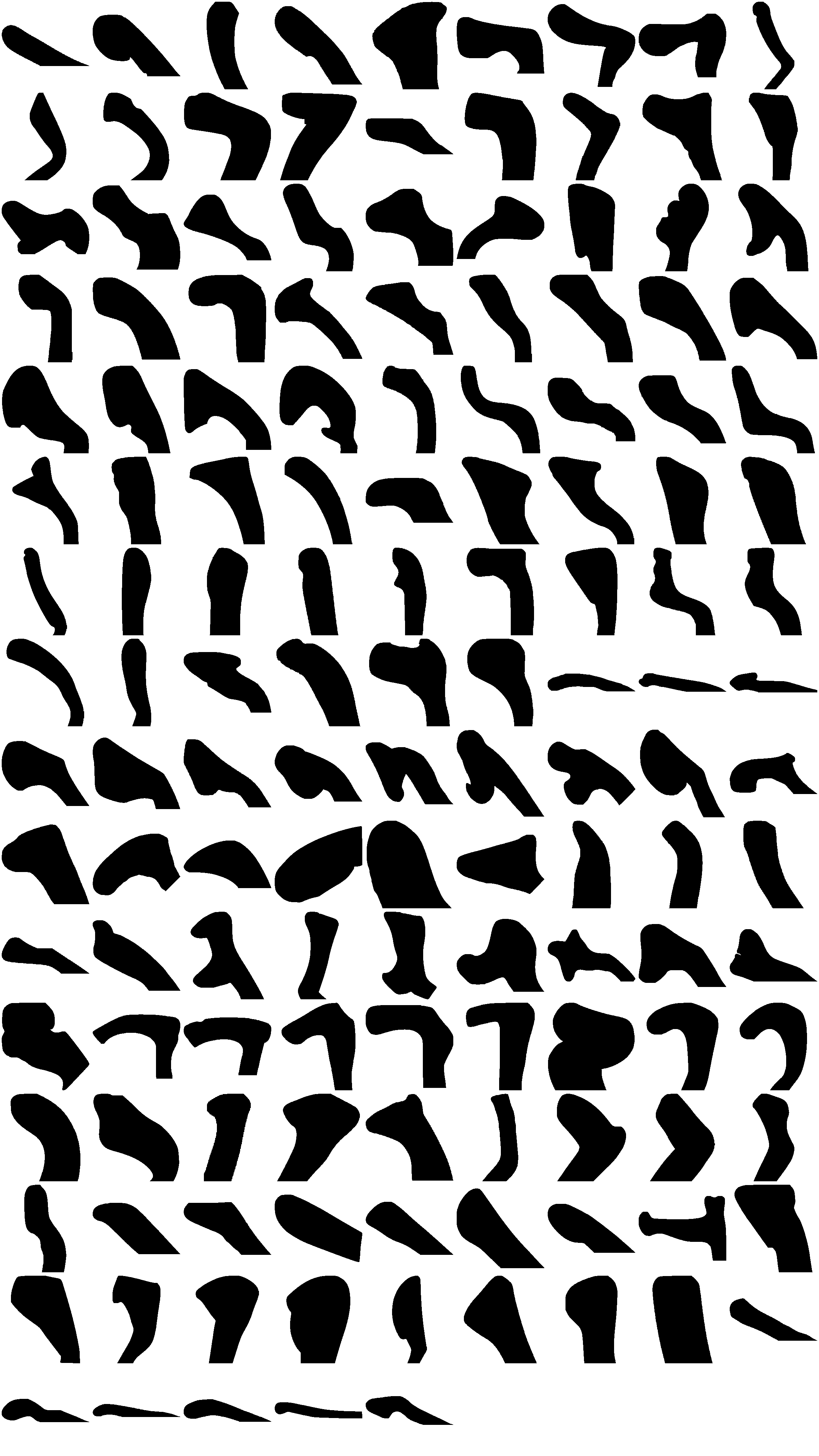}
}\quad
\subcaptionbox{OSTIA3}{
\includegraphics[height=3.8cm]{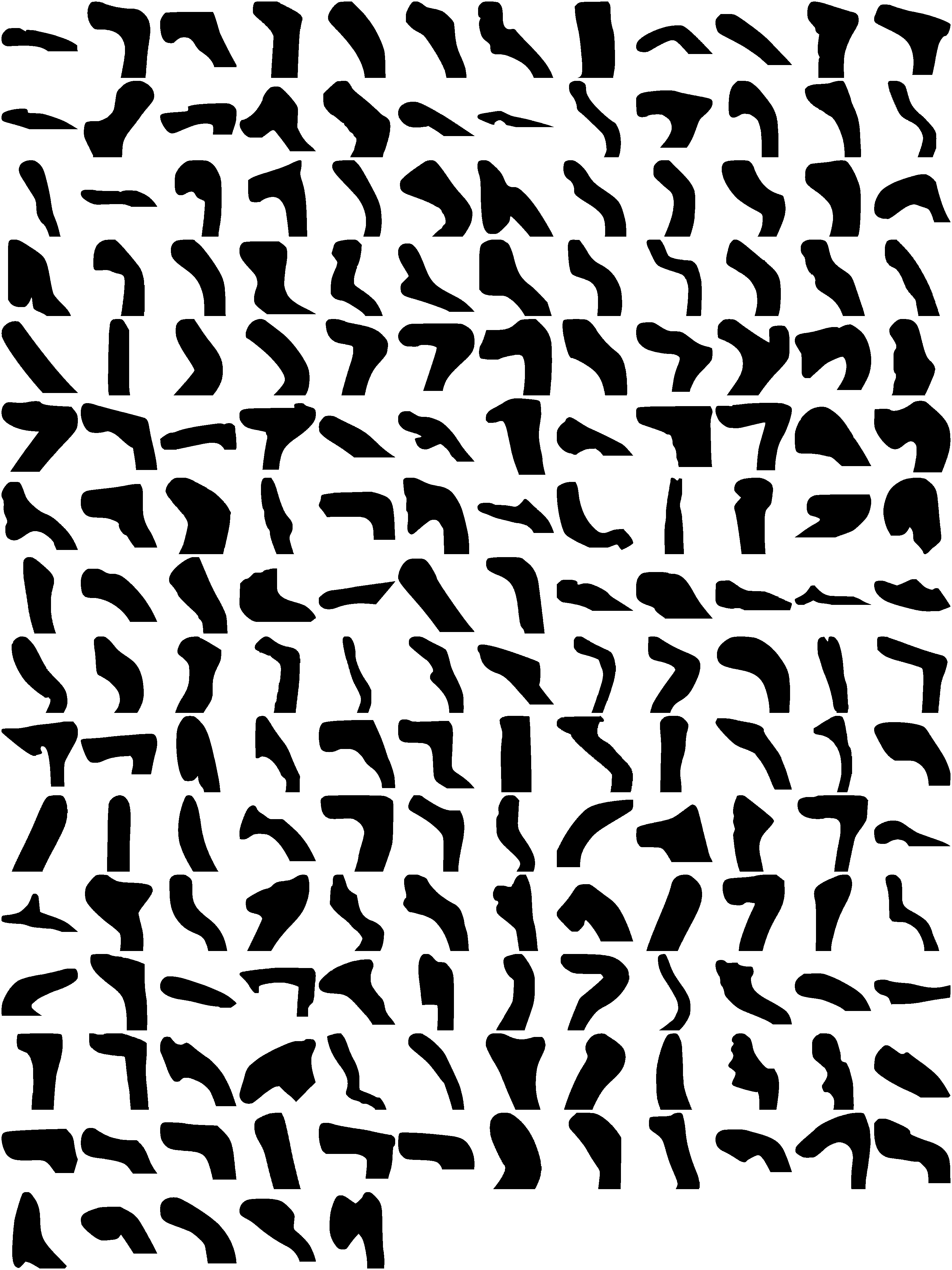}
}\quad
\subcaptionbox{OSTIA4}{
\includegraphics[height=3.8cm]{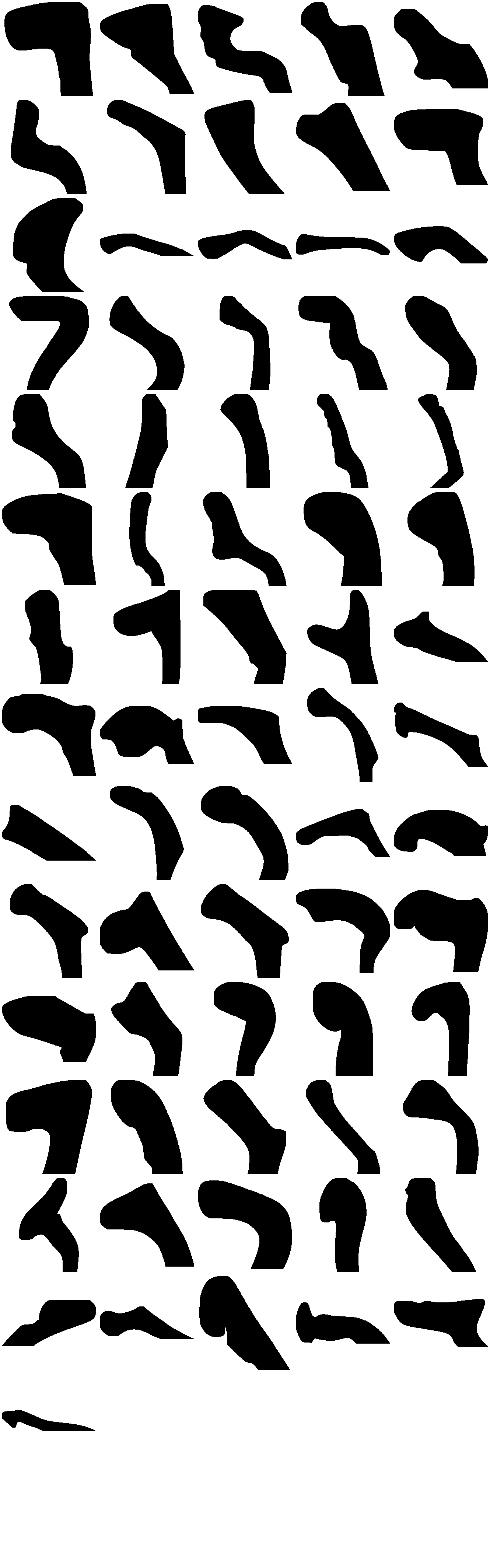}
}\quad
\subcaptionbox{PAP85}{
\includegraphics[height=3.8cm]{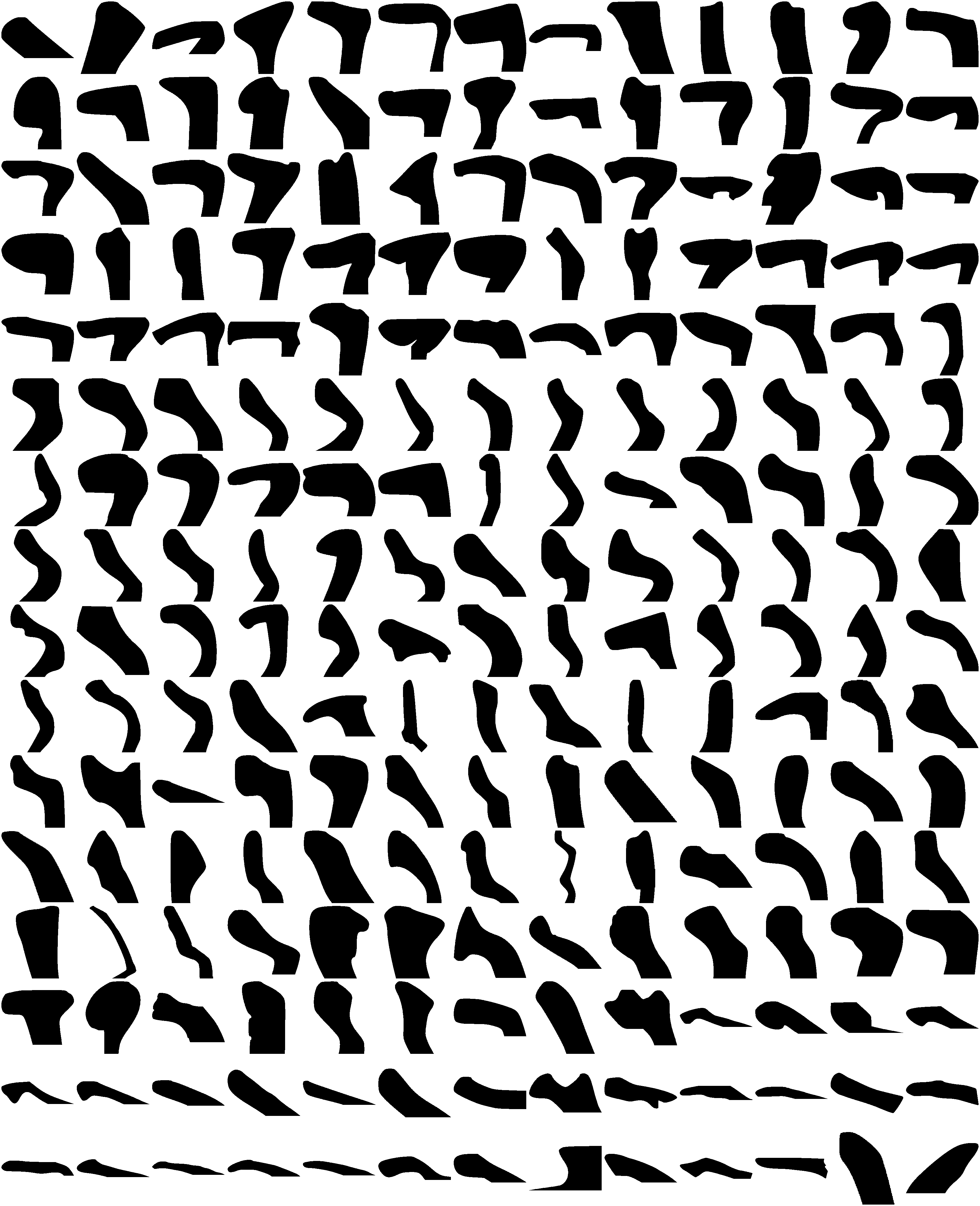}
}\quad
\subcaptionbox{POHL70}{
\includegraphics[height=3.8cm]{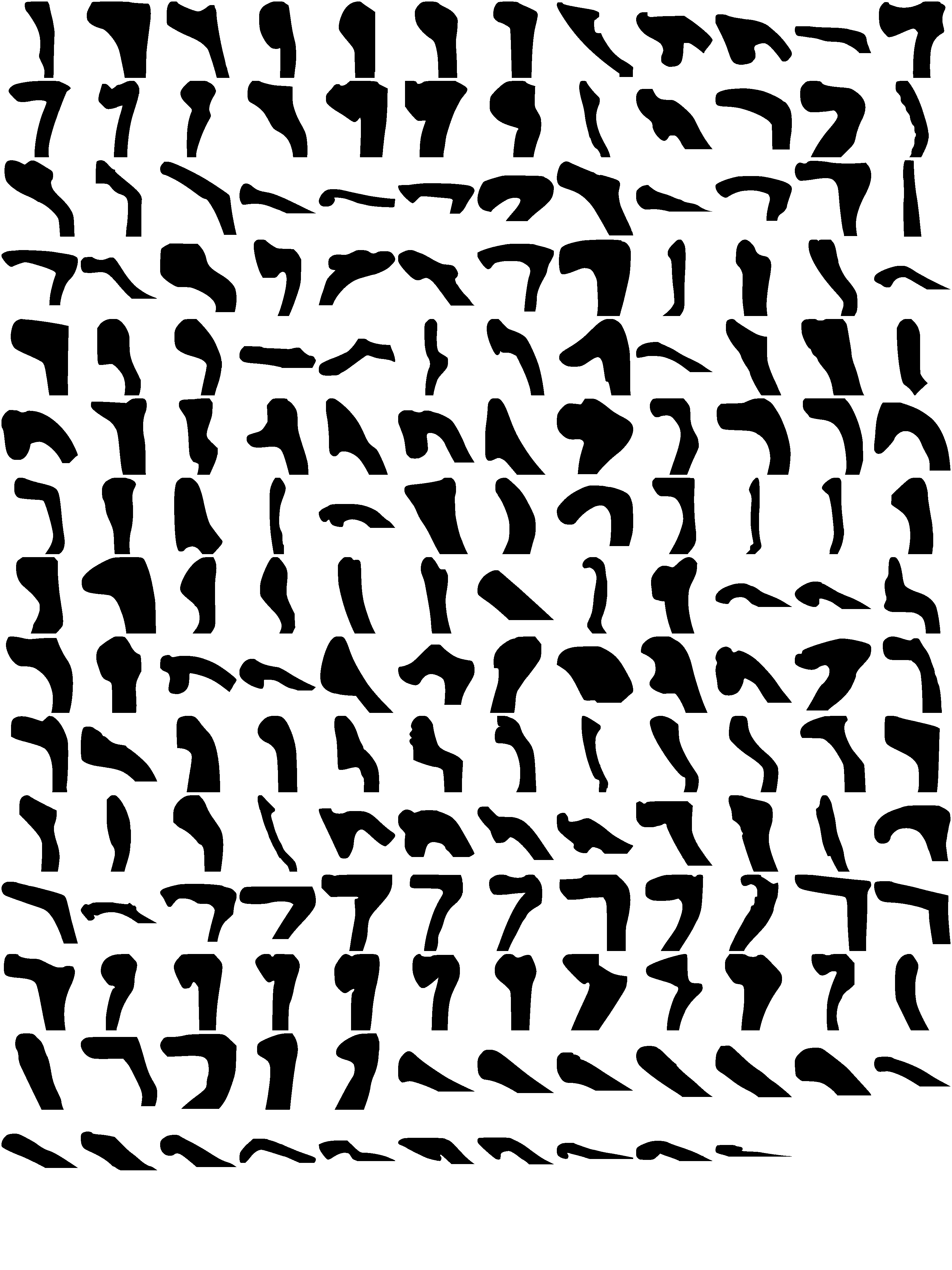}
}\quad
\subcaptionbox{ROB97}{
\includegraphics[height=3.8cm]{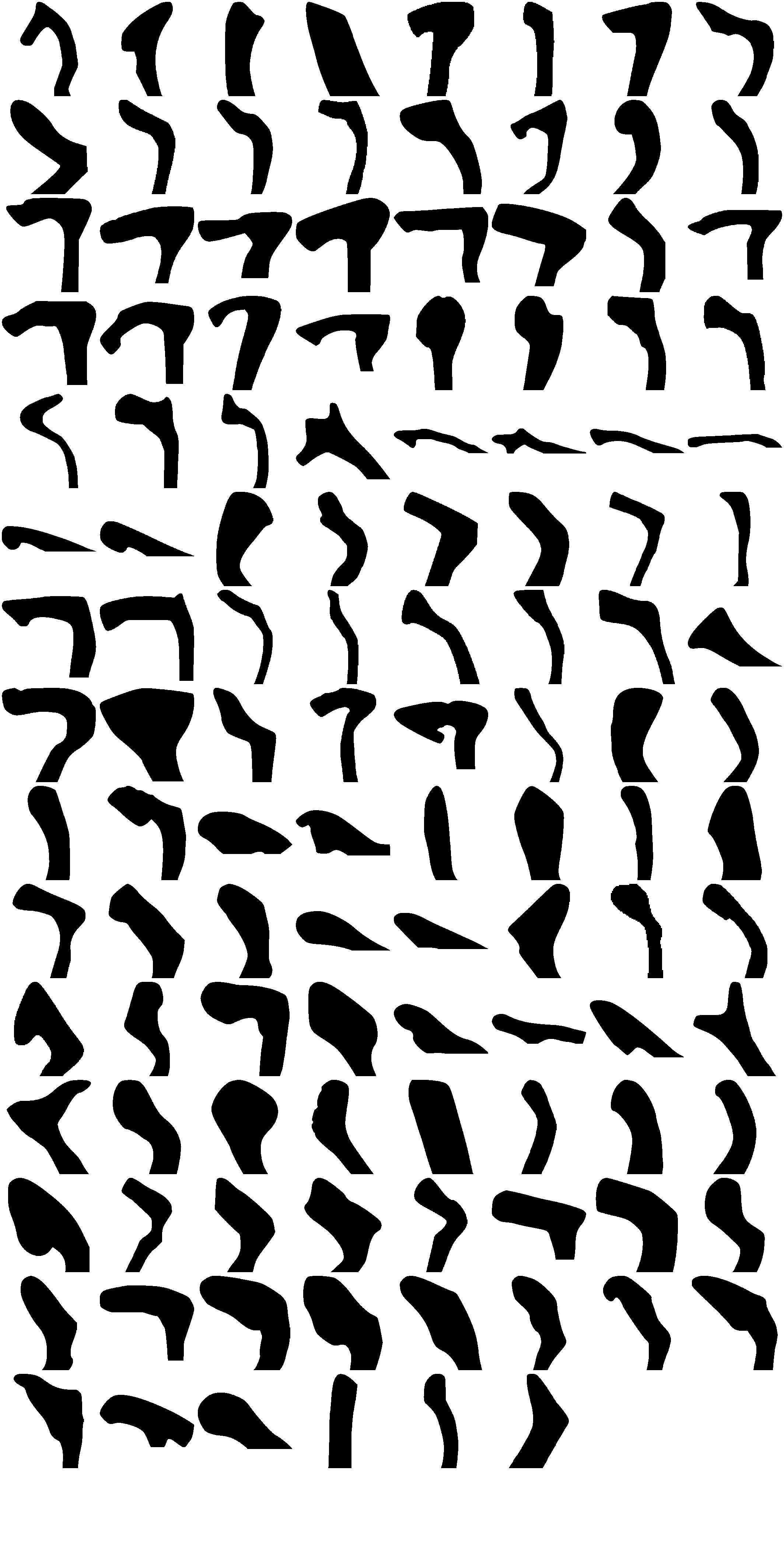}
}\quad
\subcaptionbox{SCAHO96}{
\includegraphics[height=3.8cm]{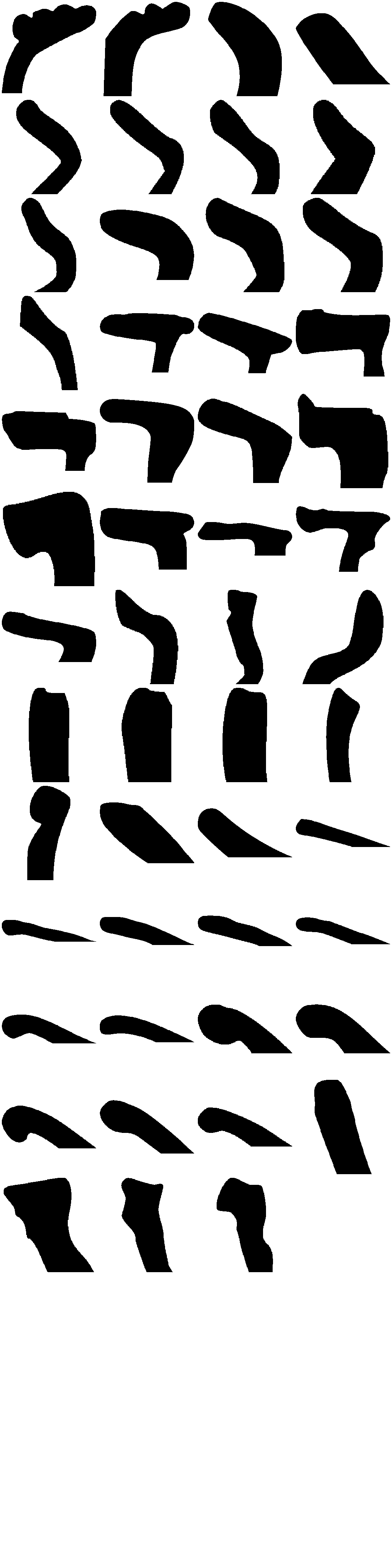}
}\quad
\subcaptionbox{STAN01}{
\includegraphics[height=3.8cm]{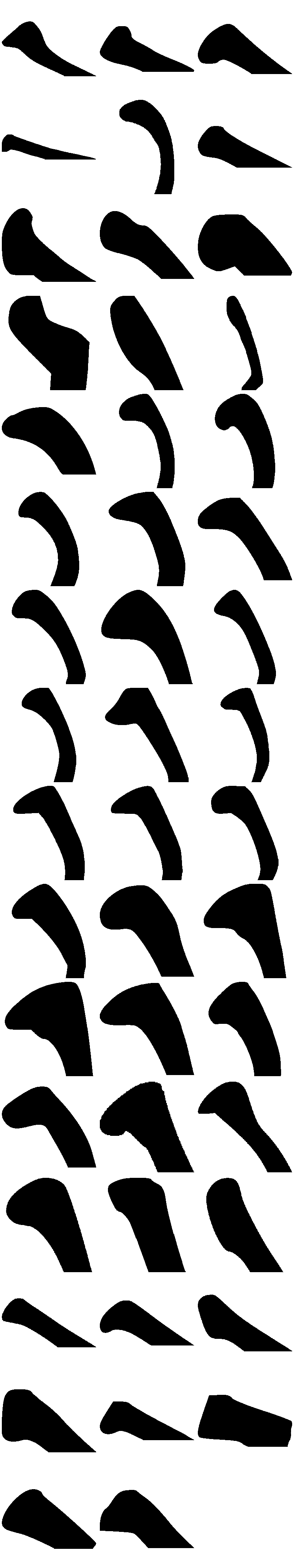}
}

    \caption{ROCOPOT database v1.0: (R)ims profiles.}
    \label{fig: database 1.0}
\end{figure*}

All the rims require a further polishing step so as to make them uniformly represented in the imaging space and before processing them into the deep learning algorithm. 
This is due to different presentation/editorial styles adopted across the corpora.
For example, in Figure \ref{fig: preprocessing} we report common situations when identifying the rim, e.g.\ its separation from the template and cutting of the elongated slope as well as mirroring (Figure \ref{fig: rim separation}), the fill in of relevant portions with a black colour with a possible rotation of the profile (Figure \ref{fig: rim filling 1}), the cleaning of scanned contours from ageing phenomena and the removal of undesired handles (Figure \ref{fig: rim polishing 1} and \ref{fig: rim polishing 2}). 
Furthermore, based on the \texttt{convert} script from Imagemagick\footnote{\url{www.imagemagick.org/script/convert.php}}, we converted the extracted rims from \texttt{.png} format to a \texttt{.svg} format so as to remove pixelisation errors (density 1200 and $1500\times 1500$ pixels).
All the rims are then resized to $256\times256$ pixels, without loosing the aspect ratio, for the purpose of data normalisation in the deep learning network described in the rest of the paper, see Figure \ref{fig: database 1.0} for a full display of rims in our database.

\begin{table*}[htb]\scriptsize
\caption{ROCOPOT database v1.0: number of (O)riginal shapes and their (B)ases, (H)andles, (R)ims and Rims with Handles (RH).}
\label{tab: database 1.0}
\begin{tabularx}{1\textwidth}{Xccc|c|c|c|ccc}
\toprule
\texttt{IDCAT} & \multirow{2}{*}{Ref.} & \multirow{2}{*}{Year} & \multicolumn{5}{c}{Shapes} & \multirow{2}{*}{Chronological range} & \multirow{2}{*}{Site} \\
\cmidrule(r){4-8}
(catalogue abbreviation) &&     &   O &  B &   H &  R &  RH &                    &\\
\midrule
BRAG96  & \cite{BRAG96}  & 1996 &  16 &   - &   1 &  16 &   2 & 200 BC   - AD 200      & Naples (Campania)\\ 
CICI96  & \cite{CICI96}  & 1996 &  50 &   8 &  12 &  42 &   3 & 21 BC   - AD 79      & Terzigno (Campania)\\
CIPR96  & \cite{CIPR96}  & 1996 &  74 &   5 &  20 &  69 &  13 & 325 BC   - AD 500      & Benevento (Samnium)\\
CM91    & \cite{CM91}    & 1991 &  43 &   4 &  10 &  39 &   4 & 100 BC - AD 200  & La Celsa (South Etruria)\\
CT84    & \cite{CT84}    & 1984 & 269 &  28 &  26 & 240 &  14 & 700 BC - AD 100  & Pompei (Campania)\\
DECA94  & \cite{DECA94}  & 1994 &  74 &   6 &  23 &  68 &   8 & 22 BC   - AD 79    & Boscoreale (Campania)\\
DIGIO96 & \cite{DIGIO96} & 1996 &  52 &   - &   3 &  52 &  17 & 325 BC   - AD 79    & Pompei (Campania)\\
DUN64   & \cite{DUN64}   & 1964 & 245 &  13 &  35 & 217 &  37 &  AD 60  - AD 70   & Sutri (South Etruria)\\
DUN65   & \cite{DUN65}   & 1965 & 127 &  13 &  11 & 107 &   9 & 150 BC -  1 BC   & Sutri (South Etruria)\\
DYS76   & \cite{DYS76}   & 1976 & 814 & 110 &  62 & 678 &  46 & 350 BC - AD 335  & Cosa (Central Etruria)\\
FEDE96  & \cite{FEDE96}  & 1996 &  97 &   9 &   4 &  88 &   7 & 200 BC   - AD 200    & Circello (Samnium)\\
FULF84  & \cite{FULF84}  & 1984 &  67 &   4 &   3 &  63 &  15 & AD 301   - AD 700    & Carthage (North Africa, imports from Italy)\\
FULF94  & \cite{FULF94}  & 1994 &  23 &   - &   - &  23 &   - & AD 25   - AD 550    & Carthage (North Africa, imports from Italy)\\
GASP96  & \cite{GASP96}  & 1996 &  70 &   2 &  38 &  60 &  19 & 200 BC   - AD 79    & Pompei (Campania)\\
LUNI2   & \cite{LUNI2}   & 1977 & 453 &  52 &  25 & 395 &  37 & 325 BC   - AD 600    & Luni (Northern Etruria)\\
OLCE93  & \cite{OLCE93}  & 1993 & 404 &   9 &  14 & 395 &  54 & 80 BC   - AD 800    & Albintimilium (Liguria)\\
OSTIA1  & \cite{OSTIA1}  & 1968 & 168 &  26 &  45 & 119 &  42 & AD 101 - AD 500  & Ostia (Latium)\\
OSTIA2  & \cite{OSTIA2}  & 1970 & 175 &  21 &  19 & 140 &  24 & AD  51 - AD 150  & Ostia (Latium)\\
OSTIA3  & \cite{OSTIA3}  & 1973 & 230 &  32 &  31 & 186 &  23 & AD  51 - AD 500  & Ostia (Latium)\\
OSTIA4  & \cite{OSTIA4}  & 1977 &  90 &  13 &   9 &  74 &  14 & AD 251 - AD 425  & Ostia (Latium)\\
PAP85   & \cite{PAP85}   & 1985 & 240 &  22 &  18 & 210 &  18 & 50 BC   - AD 600    & Settefinestre (Central Etruria)\\
POHL70  & \cite{POHL70}  & 1970 & 182 &   3 &   9 & 179 &  22 & 200 BC - AD 140  & Ostia (Latium)\\
ROB97   & \cite{ROB97}   & 1997 & 132 &  12 &  21 & 119 &  13 & AD 101 - AD 635  & Mola di Monte Gelato (South Etruria)\\
SCAHO96 & \cite{SCAHO96} & 1996 &  51 &   - &  10 &  51 &   9 & 21 BC   - AD 79    & Ercolano (Campania)\\
STAN01  & \cite{STAN01}  & 2001 &  62 &  15 &   1 &  48 &   1 & 300 BC   - AD 101    & Frassineta Franco (South Etruria)\\
\midrule
\textbf{ALL} &           &    &  4208 & 407 & 450 & 3678 & 451 & &\\ 
\bottomrule
\end{tabularx}
\end{table*}

\section{Proposed approach}\label{sec: problem}
After preprocessing the ROCOPOT database and with the \mod{rim} profiles at \mod{our} disposal, we proceed with modelling the manifold of profiles by learning non-linear \mod{features associated to them}.
Such features will be used for the hierarchical clustering of the dataset. Note that the combination of these two approaches makes our workflow totally unsupervised, a key assumption for unveiling hidden or new clustering patterns in existing archaeological corpora.

\subsection{Variational Autoencoders.}
Our workflow is based on the learning of probabilistic features associated with profiles, able to ease the profile matching problem in a low-dimensional feature space.
Instead of fixing a-priori qualitative profile features (e.g.\ the diameter, the ballness, the elongation, the area, the perimeter and many more), we fix the dimension $k>0$ of the latent space and maximise the probability that profiles, generated from a stochastic sampling process in the latent space, match with the profiles in our database. 
In this sense, deep learning convolutional Variational Autoencoders (VAEs) are powerful tools that can be interpreted as a generative non-linear version of the principle component analysis (PCA).

Vanilla convolutional \emph{Autoencoders} (AEs) are artificial neural networks designed for replicating the input at its output by means of back-propagating the reconstructed result with a single layer strategy \cite{ng2011sparse}.
Here, the representation of the input is learned via the minimisation of a \emph{loss} function, composed by a reconstruction error measure and a regulariser, in a low-dimensional latent space. 
Any AE is composed by an \emph{encoder}, which reduces the dimension of the input to the latent space, and a \emph{decoder}, which reconstructs the output from the latent space.
More advanced \emph{Stacked Sparse Autoencoders} (SSAEs) \cite{VicLarLajBenMan2010} networks concatenate multiple vanilla AE, where each latent space is the input of the subsequent AE. \mod{Here, the sparsity assumption forces the neurons to specialise on few high-level features} \cite{Xu2016}.
However, both AE and SSAE are limited by their (possibly) discontinuous latent space, making thus impossible the interpolation between multiple input data and leading to the \emph{overfitting} phenomenon. This is a limitation for scaling-up the approach when processing unseen data. 

In contrast, \emph{Variational Autoencoders} (VAEs) \cite{KinWel2014,Higgins2017} are generative deep learning convolutional neural networks able to learn a continuous probabilistic representation of the global potsherds features in the latent space. 
The loss function in the network is composed by a regulariser term plus a reconstruction term:
\begin{equation}
\mathcal{L}(\xbold,\overline{\xbold})=
\beta_e\mathcal{L}_{KL}(\xbold,\zbold) 
+
\mathcal{L}_{c}(\xbold,\overline{\xbold}),
\label{eq: main loss}
\end{equation}
where $\beta_e>0$ is a weighting parameter varying according to the learning epoch $e$, \mod{$\mathcal{L}_{KL}(\xbold,\zbold)$ is the \mod{loss function} accounting for the reverse Kullback-Leibler (KL) divergence $D_{\text{KL}}$ on the latent space, see \eqref{eq:  DKL rearranged} in the Appendix \ref{sec: appendix} and Figure \ref{fig: VAE network}, 
while the $\mathcal{L}_{c}(\xbold,\overline{\xbold})$ is a sigmoid cross-entropy energy with logits loss, see \eqref{eq: reconstructionloss} in Appendix \ref{sec: appendix}.}
By acting as regulariser, $\mathcal{L}_{KL}(\xbold,\zbold)$ ensures that the potsherds distribution generated from the learned latent space is as close as possible to the target one, while $\mathcal{L}_{c}(\xbold,\overline{\xbold})$ measures the probability error of reconstructing black-white pixels in a potsherd image, similarly to a labelling task.
More details about our chosen structure of the VAE network and technical parameters for the training process are given in Section \ref{sec: results}. 

\subsection{Hierarchical clustering.}\label{sec: hcluster}
Once the VAE network is trained, we can associate to each profile a low-dimentional latent vector $\zbold\in\RR^{k}$ with $k>0$. 
From the reparametrisation trick discussed in \eqref{eq: reparametrisation} in Section \ref{sec: reparametrisation}, i.e.\ $\zbold=\bm{\mu}+\bm{\sigma}\odot\bm{\varepsilon}$ where $\bm{\varepsilon}\in\RR^k$ is a random vector, we concatenate the mean $\bm{\mu}\in\RR^k$ and the variance $\bm{\sigma}\in\RR^k$ in a vector of features $\overline{\zbold}\in\RR^{2k}$.
These features \mod{can} now be used as a shape \emph{signature} for the clustering task.
Since popular clustering methods, like \texttt{k-means}, have the bottleneck of requiring as input the number of expected clusters (unknown for our database), we opted for an \emph{agglomerative hierarchical clustering} approach so as to build a tree of nested clusters. 
\mod{Starting} with a single class for each potsherd profile, we subsequently merge together profiles sharing similar features in a bottom-up hierarchical strategy according to a \texttt{method} for computing the distance between clusters and a \texttt{metric} between the observed features. In this way, we can inspect different levels of similarities and shape relations.
In order to automatically identify the best cluster merging rule, we compute the cophenetic coefficient score \cite{SokRoh1962} for each pair of \texttt{methods} (nearest distance \cite{Florek1951}, farthest distance \cite{sorensen1948method}, UPGMA  \cite{sokal58}, WPGMA \cite{sokal58}, UPGMC \cite{Milligan1980}, WPGMC \cite{JainDubes1988} and ward \cite{Ward1963}) and \texttt{metrics} (Euclidean, cityblock, Chebychev and cosine). Here, the cophenetic score \mod{\cite{Clarke2016}} measures how faithfully the selected tree represents the dissimilarities among observation and the pair returning the highest value is selected as the best pair. 

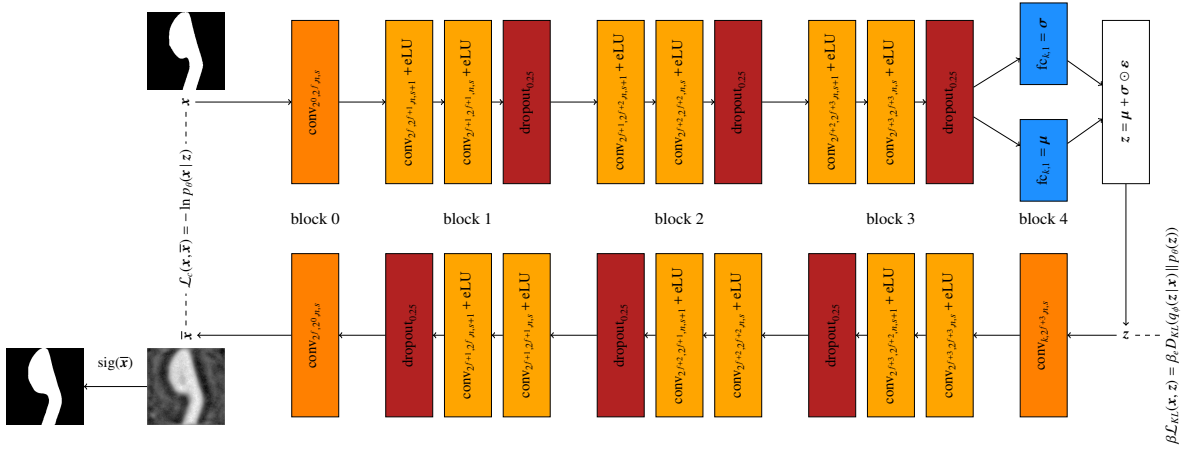
\begin{figure*}[!htb]\centering
    \resizebox{0.95\textwidth}{!}{
    \begin{tikzpicture}
    \node[inner sep=0pt] (whitehead1) at (0,1.1)
    {\includegraphics[width=.1\textwidth]{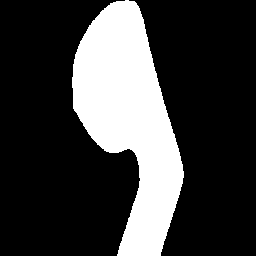}};
    \node[inner sep=0pt] (whitehead2) at (0,-6.1)
    {\includegraphics[width=.1\textwidth]{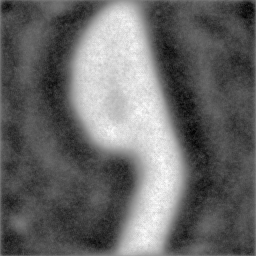}};
    \node[inner sep=0pt] (whitehead3) at (-3,-6.1)
    {\includegraphics[width=.1\textwidth]{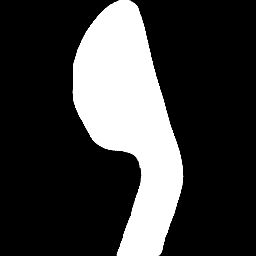}};
    \node (sigx) at (-1.5,-5.6) {\small$\sigmoid(\overline{\xbold})$};
    \node[rotate=90] (y) at (0,0) {\small$\xbold$};
    \node[convzero,rotate=90,minimum width=3.5cm] (conv0) at (2.75,0) {\small$\text{conv}_{2^0,2^f,n,s}$};
    \node[conv,rotate=90,minimum width=3.5cm] (conv11) at (4.75,0) {\small$\text{conv}_{2^f,2^{f+1},n,s+1}$\,+\,$\eLU$};
    \node[conv,rotate=90,minimum width=3.5cm] (conv12) at (6.00,0) {\small$\text{conv}_{2^{f+1},2^{f+1},n,s}$\,+\,$\eLU$};
    \node[drop,rotate=90,minimum width=3.5cm] (drop1) at (7.25,0) {\small$\text{dropout}_{0.25}$};
    \node[conv,rotate=90,minimum width=3.5cm] (conv21) at (9.25,0) {\small$\text{conv}_{2^{f+1},2^{f+2},n,s+1}$\,+\,$\eLU$};
    \node[conv,rotate=90,minimum width=3.5cm] (conv22) at (10.50,0) {\small$\text{conv}_{2^{f+2},2^{f+2},n,s}$\,+\,$\eLU$};
    \node[drop,rotate=90,minimum width=3.5cm] (drop2) at (11.75,0) {\small$\text{dropout}_{0.25}$};
    \node[conv,rotate=90,minimum width=3.5cm] (conv31) at (13.75,0) {\small$\text{conv}_{2^{f+2},2^{f+3},n,s+1}$\,+\,$\eLU$};
    \node[conv,rotate=90,minimum width=3.5cm] (conv32) at (15.00,0) {\small$\text{conv}_{2^{f+3},2^{f+3},n,s}$\,+\,$\eLU$};
    \node[drop,rotate=90,minimum width=3.5cm] (drop3) at (16.25,0) {\small$\text{dropout}_{0.25}$};
    \node[fc,rotate=90, minimum width = 1.75cm] (fc51) at (18.25,1.25) {\small$\text{fc}_{k,1}=\bm{\sigma}$};
    \node[fc,rotate=90, minimum width = 1.75cm] (fc52) at (18.25,-1.25) {\small$\text{fc}_{k,1}=\bm{\mu}$};
    \node[view,rotate=90,minimum width=3.5cm] (kld) at (20.00,0) {\small $\zbold=\bm{\mu}+\bm{\sigma}\odot\bm{\varepsilon}$};
    \node (z) at (20.00,-5){\small$\zbold$};
    \node[convzero,rotate=90,minimum width=3.5cm] (tconv51) at (18.25,-5) {\small$\text{conv}_{k,2^{f+3},n,s}$};
    \node[conv,rotate=90,minimum width=3.5cm] (tconv32) at (16.25,-5) {\small$\text{conv}_{2^{f+3},2^{f+3},n,s}$\,+\,$\eLU$};
    \node[conv,rotate=90,minimum width=3.5cm] (tconv31) at (15.00,-5) {\small$\text{conv}_{2^{f+3},2^{f+2},n,s+1}$\,+\,$\eLU$};
    \node[drop,rotate=90,minimum width=3.5cm] (tdrop3) at (13.75,-5) {\small$\text{dropout}_{0.25}$};
    \node[conv,rotate=90,minimum width=3.5cm] (tconv22) at (11.75,-5) {\small$\text{conv}_{2^{f+2},2^{f+2},n,s}$\,+\,$\eLU$};
    \node[conv,rotate=90,minimum width=3.5cm] (tconv21) at (10.50,-5) {\small$\text{conv}_{2^{f+2},2^{f+1},n,s+1}$\,+\,$\eLU$};
    \node[drop,rotate=90,minimum width=3.5cm] (tdrop2) at (09.25,-5) {\small$\text{dropout}_{0.25}$};
    \node[conv,rotate=90,minimum width=3.5cm] (tconv12) at (07.25,-5) {\small$\text{conv}_{2^{f+1},2^{f+1},n,s}$\,+\,$\eLU$};
    \node[conv,rotate=90,minimum width=3.5cm] (tconv11) at (06.00,-5) {\small$\text{conv}_{2^{f+1},2^{f},n,s+1}$\,+\,$\eLU$};
    \node[drop,rotate=90,minimum width=3.5cm] (tdrop1) at (04.75,-5) {\small$\text{dropout}_{0.25}$};
    \node[convzero,rotate=90,minimum width=3.5cm] (tconv00) at (02.75,-5) {\small$\text{conv}_{2^{f},2^{0},n,s}$};
    \node[rotate=90] (ry) at (0,-5) {\small$\overline{\xbold}$};
    \node[rotate=90] (L) at (0, -2.5) {\small$\mathcal{L}_{c}(\xbold,\overline{\xbold}) = -\ln p_\theta( \xbold\,|\,\zbold)$};
    \node[rotate=90] (KLD) at (21, -5) {\small$\beta\mathcal{L}_{KL}(\xbold,\zbold) = \beta_e D_{KL}(q_\phi(\zbold\,|\,\xbold)\,\|\, p_\theta(\zbold))$};
    \draw[->] (y) -- (conv0);
    \draw[->] (conv0) -- (conv11);
    \draw[->] (conv11) -- (conv12);
    \draw[->] (conv12) -- (drop1);
    \draw[->] (drop1) -- (conv21);
    \draw[->] (conv21) -- (conv22);
    \draw[->] (conv22) -- (drop2);
    \draw[->] (drop2) -- (conv31);
    \draw[->] (conv31) -- (conv32);
    \draw[->] (conv32) -- (drop3);
    \node (b0) at (2.75, -2.5) {block 0};
    \node (b1) at (6, -2.5) {block 1};
    \node (b2) at (10.5, -2.5) {block 2};
    \node (b3) at (15, -2.5) {block 3};
    \node (b4) at (18.25, -2.5) {block 4};
    \draw[->] (drop3) -- (fc51);
    \draw[->] (drop3) -- (fc52);
    \draw[->] (fc51) -- (kld);
    \draw[->] (fc52) -- (kld);
    \draw[->] (kld) -- (z);
    \draw[-,dashed] (KLD) -- (z);
    \draw[->] (z) -- (tconv51);
    \draw[-,dashed] (y) -- (L);
    \draw[-,dashed] (ry) -- (L);
    \draw[->] (tconv51) -- (tconv32);
    \draw[->] (tconv32) -- (tconv31);
    \draw[->] (tconv31) -- (tdrop3);
    \draw[->] (tdrop3) -- (tconv22);
    \draw[->] (tconv22) -- (tconv21);
    \draw[->] (tconv21) -- (tdrop2);
    \draw[->] (tdrop2) -- (tconv12);
    \draw[->] (tconv12) -- (tconv11);
    \draw[->] (tconv11) -- (tdrop1);
    \draw[->] (tdrop1) -- (tconv00);
    \draw[->] (tconv00) -- (ry);
     \draw[->] (whitehead2) -- (whitehead3);
    \end{tikzpicture}
    }
    \caption{Proposed Variational Autoencoder network. Note that the negative image of the potsherds is used for a better processing.}
    \label{fig: VAE network}
    \end{figure*}
    
\section{Network Parameters and Results}\label{sec: results}
In this section we further discuss the parameters of our VAE network and the results produced by our workflow, when applied to the rims in our database.
All the tests are performed on a GPU Nvidia Quadro P6000. The code is implemented in MATLAB 2020b.

\subsection{Parameters and analysis of the VAE network.}

Our VAE network is composed by multiple convolutional layers of filter size $3\times3$, concatenated with ad-hoc exponential linear unit (eLU) and dropout layers\footnote{dropout layers are disable during the inference step.} (with fixed dropout parameter equal to $0.25$). With reference to Figure \ref{fig: VAE network} where the VAE network used in this work is depicted, $\xbold$ is the original black-white profile ($256\times 256$ pixels) and $\overline{\xbold}$ is the output of the network. Note that $\overline{\xbold}$ is required for computing the reconstruction loss $\mathcal{L}_c$ in \eqref{eq: main loss}, while the output profile is obtained by $\sigmoid(\overline{\xbold})$. Also, each convolutional layer is denoted by $\mathrm{conv}_{f_1,f_2,n,s}$, where $f_1$ are the number of input channels, $f_2$ are the number of filters, $n$ is the spatial size of the filters and $s$ is the stride. In the network, parameters are set to $f=4$, $n=3$ and $s=1$ while $k=128$ is the dimension of the latent space.

We account for handcrafted details, printing or scanning artefacts by increasing the number of potsherds at our disposal \mod{$5$ times} via data augmentation, e.g.\ by adding variations of the original shapes obtained via image operations like erosion and dilation (of $3$ pixels) and rotations (by $-5$ and $+5$ degrees).
We trained our network on 90\% of the rims available, using the remaining 10\% as test data, and for a maximum number of 1000 epochs. 
The learned parameters are updated using the Adaptive Moment Estimation (ADAM) \cite{ADAM}, with constant learning rate of 1\eu-06.

\begin{figure}[htb]
    \centering
    \hfill
    \begin{subfigure}[t]{0.465\textwidth}\centering
    \includegraphics[width=0.965\textwidth]{{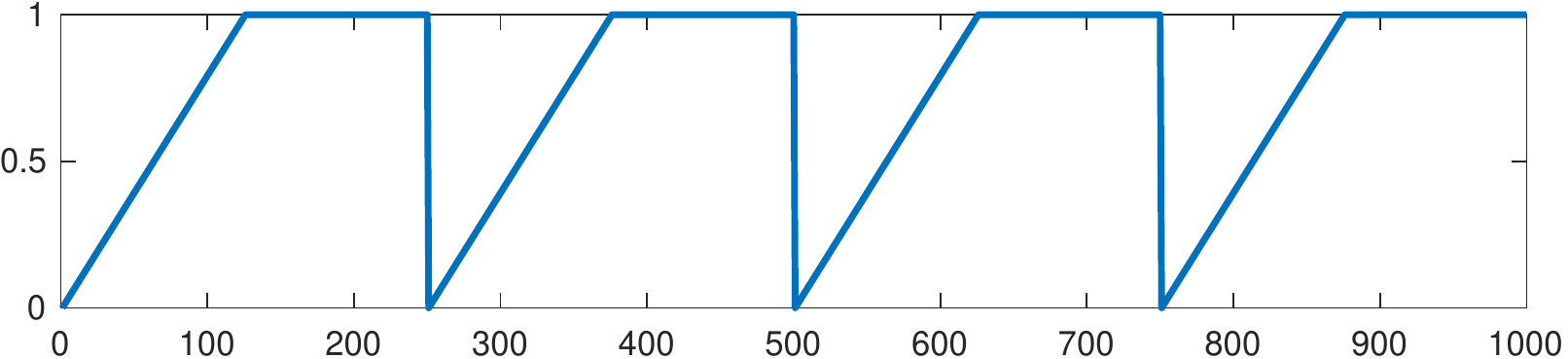}}    
    \caption{Cyclical annealing of $\beta_e$ ($T$=1000, $M$=4, $R$=0.5).}
    \label{fig: annealing}
    \end{subfigure}
    \hfill
    \begin{subfigure}[t]{0.465\textwidth}\centering
    \includegraphics[width=1\textwidth]{{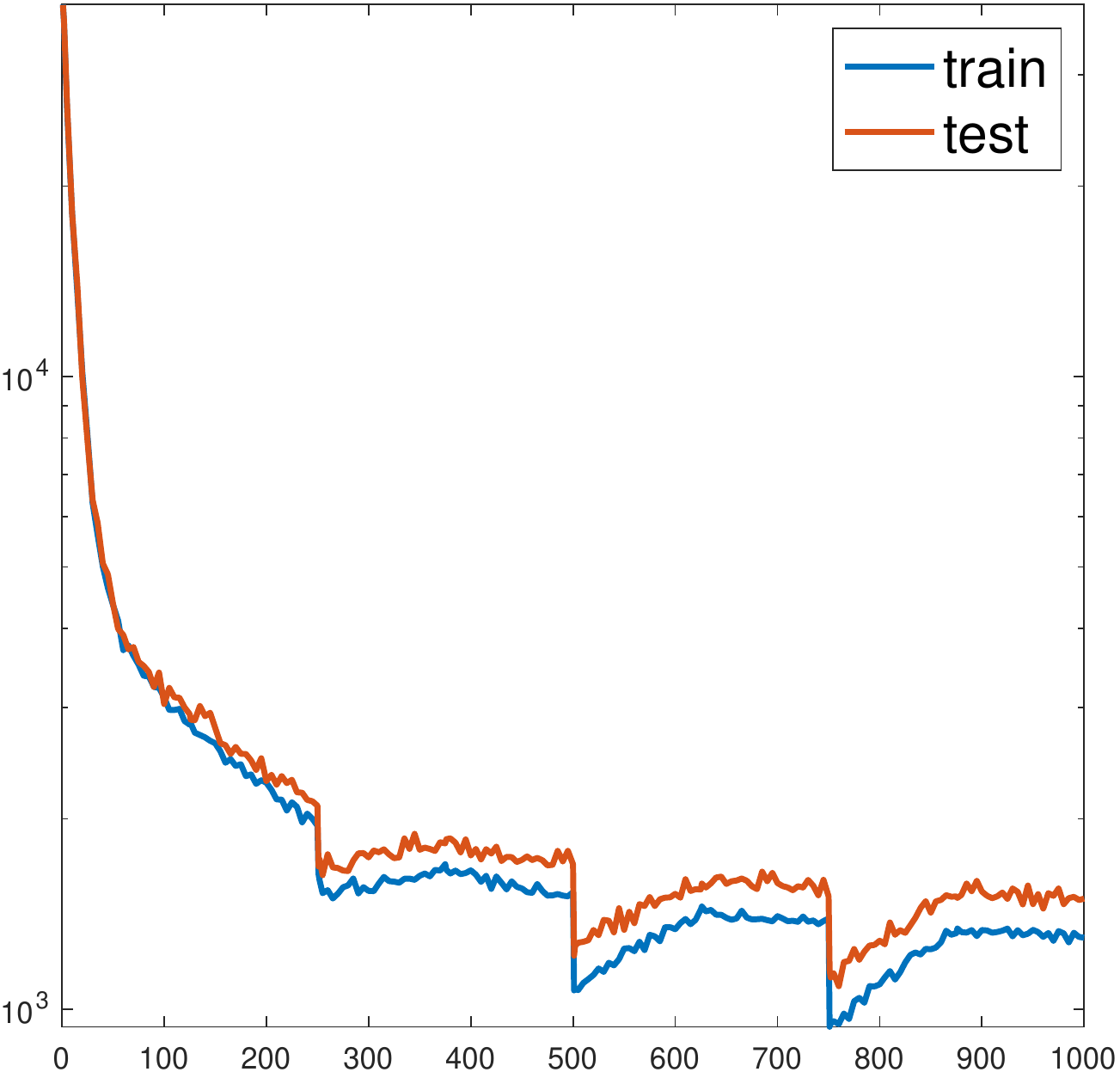}}
    \caption{ELBO loss in semi-logarithm scale along $y$-coordinate.}
    \label{fig: elbo loss semilogy}
    \end{subfigure}
    \caption{From top to bottom: Cosine decay of the learning rate;
    cyclical annealing of $\beta_e$ and ELBO loss \eqref{eq: main loss} (train in blue and and test in red, computed on randomly selected subsamples of 32 shapes, repeated 10 times and averaged).}
\end{figure}

In order to avoid the so-called \emph{KL-vanishing} phenomenon, related to the undesired vanishing of the regulariser during the optimisation, the regularisation term is weighted by a positive factor $\beta_e\in(0,1]$. Among different choices \cite{Bowman2016,Zhao2017} (applied to natural language processing challenges), we follow the cyclical annealing approach \cite{Fu2019} with a periodical adjustment of the weight $\beta_e$ of \eqref{eq: main loss} in $M$ cycles. 
For a fixed epoch $e$, the weight $\beta_e$ is computed according to the following rule:
\[
\beta_e
=
\begin{dcases}
f(\tau_e)&\text{if }\tau\leq R,\\
1& \text{otherwise}
\end{dcases}
\]
for a linear function $f(\tau_e)=\tau_e/R$ and 
\[
\tau_e=\mathrm{mod}(e-1,\ceil{T/M})/(T/M),
\] 
\begin{figure}[!htb]
    \centering
    \begin{subfigure}[t]{0.475\textwidth}
    \includegraphics[width=1\textwidth,trim=0cm 1.635cm 0cm 0cm,clip=true]{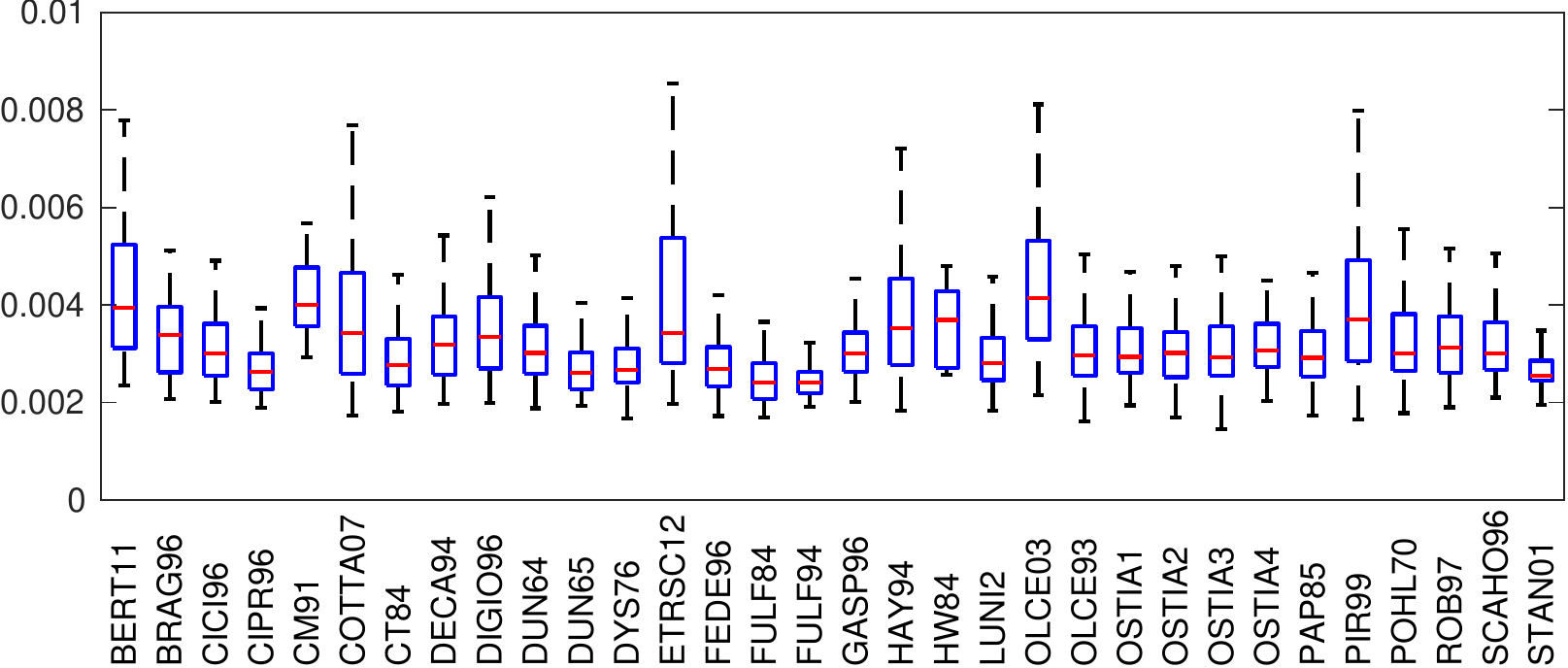}
    \end{subfigure}
    \begin{subfigure}[t]{0.475\textwidth}
    \includegraphics[width=1\textwidth]{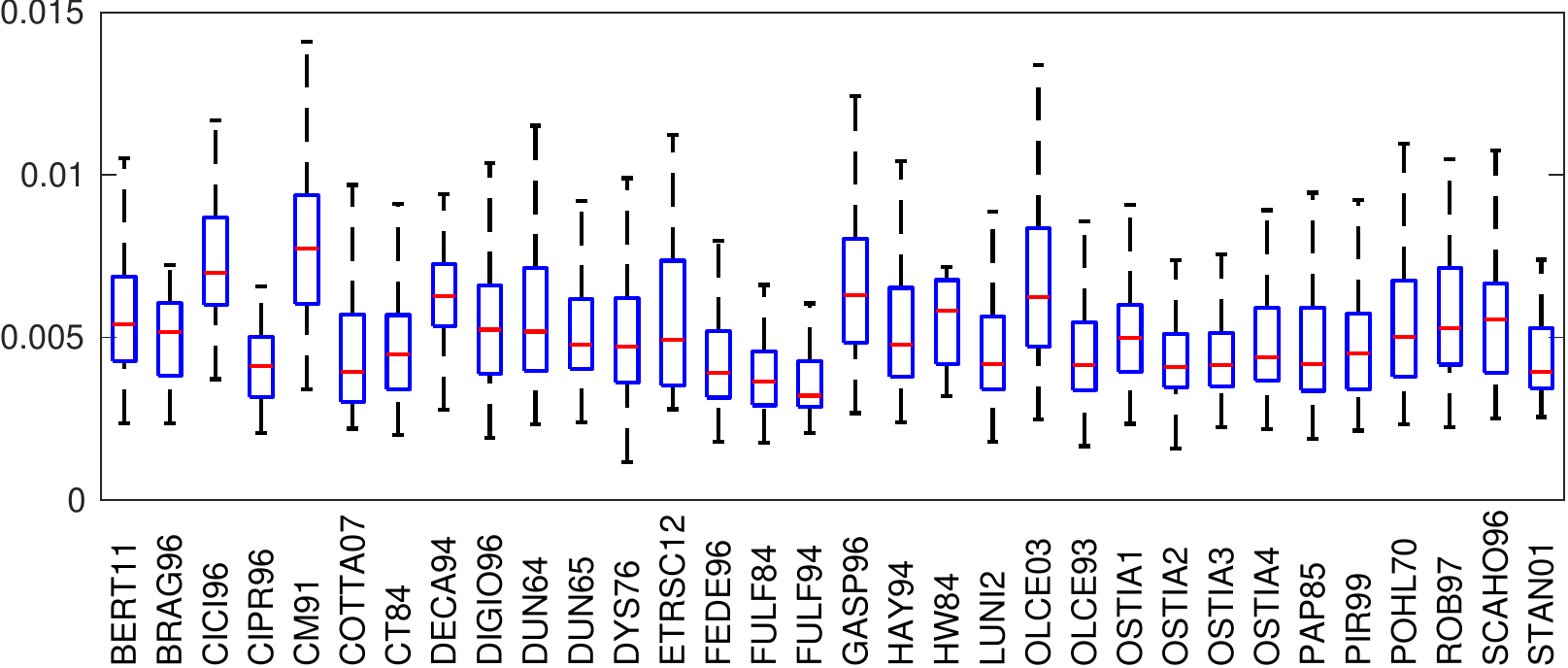}
    \end{subfigure}
    \caption{Boxplots: MSE (top) and IoU (bottom) errors.}
    \label{fig: boxplot}
\end{figure}
where $T$ is maximum number of epochs, $M$ is the number of cycles and $R$ is the \mod{rate of which} to increase $\beta_e$ within a cycle. 
Thus, each cycle is composed by an \emph{annealing stage} where $\beta_e$ linearly increases from 0 to 1, and a \emph{fixing} stage when $\beta_e=1$. This strategy forces the model to behaves like AEs when starting each cycle with $\beta_e=0$ 
and like standard VAEs when $\beta_e=1$; in contrast, the model transits from a point estimate to a distribution estimate when linearly increasing $\beta_e$ from 0 to 1.
The cyclic repetition guarantees a progressive learning, leading to a more meaningful latent space.
In our experiments, we fixed $T=1000$, $M=4$ and $R=0.5$, for a cycling annealing strategy as depicted in Figure \ref{fig: annealing}.
Note that if \mod{$\beta_e\gg 1$} the loss would have related to the $\beta$-VAE extension, proposed by \cite{Higgins2017} and used for learning disentangled representation features in the latent space \cite{Bengio2013}, but at the price of \mod{a reduced} reconstruction quality \cite{Shao2020}.

Once our network is trained we are able to model \mod{the} probability distribution of potsherds with $\zbold\in\RR^k$ ($k=128$ parameters), and to reconstruct the ROCOPOT dataset from Figure \ref{fig: database 1.0}.
We report in Figure \ref{fig: boxplot} the confidence intervals of the pixel-wise mean squared error and the intersection over \mod{the} union error as a distance \cite{IoUloss},  defined for a shape $s$ respectively as:
\[
\text{MSE}(\xbold_s,\sigmoid(\overline{\xbold}_s)) 
= 
\frac{\sum_{i,j=1}^{H,W}(\xbold_s-\sigmoid(\overline{\xbold}_s)))^2}{H\cdot W},
\]
where $H,W$ are the dimension of the image, $\sigmoid$ is the sigmoid function, and
\[
\text{IoU}(\xbold_s,\sigmoid(\overline{\xbold}_s))
= 
1-\frac{2\sum_{i,j=1}^{H,W} (\xbold_s \cdot \sigmoid(\overline{\xbold}_s))}{\sum_{i,j=1}^{H,W} (\xbold_s)^2 + \sum_{i,j=1}^{H,W} \sigmoid(\overline{\xbold}_s)^2}.
\]

In Figure \ref{fig:  elbo loss semilogy} we display the loss function on both train and test data, individual terms (KL loss and cross-entropy reconstruction loss) and the \mod{Euclidean} error through epochs, showing no overfitting during the training of the network.
Both visual and quantitative results confirms that the shapes are reconstructed almost correctly, meaning that the neurons in the latent space are now specialised in modelling the probability distribution of the potsherds in our database.

Before using the $2k(=256)$ features for the hierarchical clustering step, a natural question is to investigate the quality of the reconstruction and the robustness of the latent space to some perturbations.

\subsection{Robustness of the learned features.}
\begin{figure*}[!htb]
    \centering
    \includegraphics[width=0.31\textwidth]{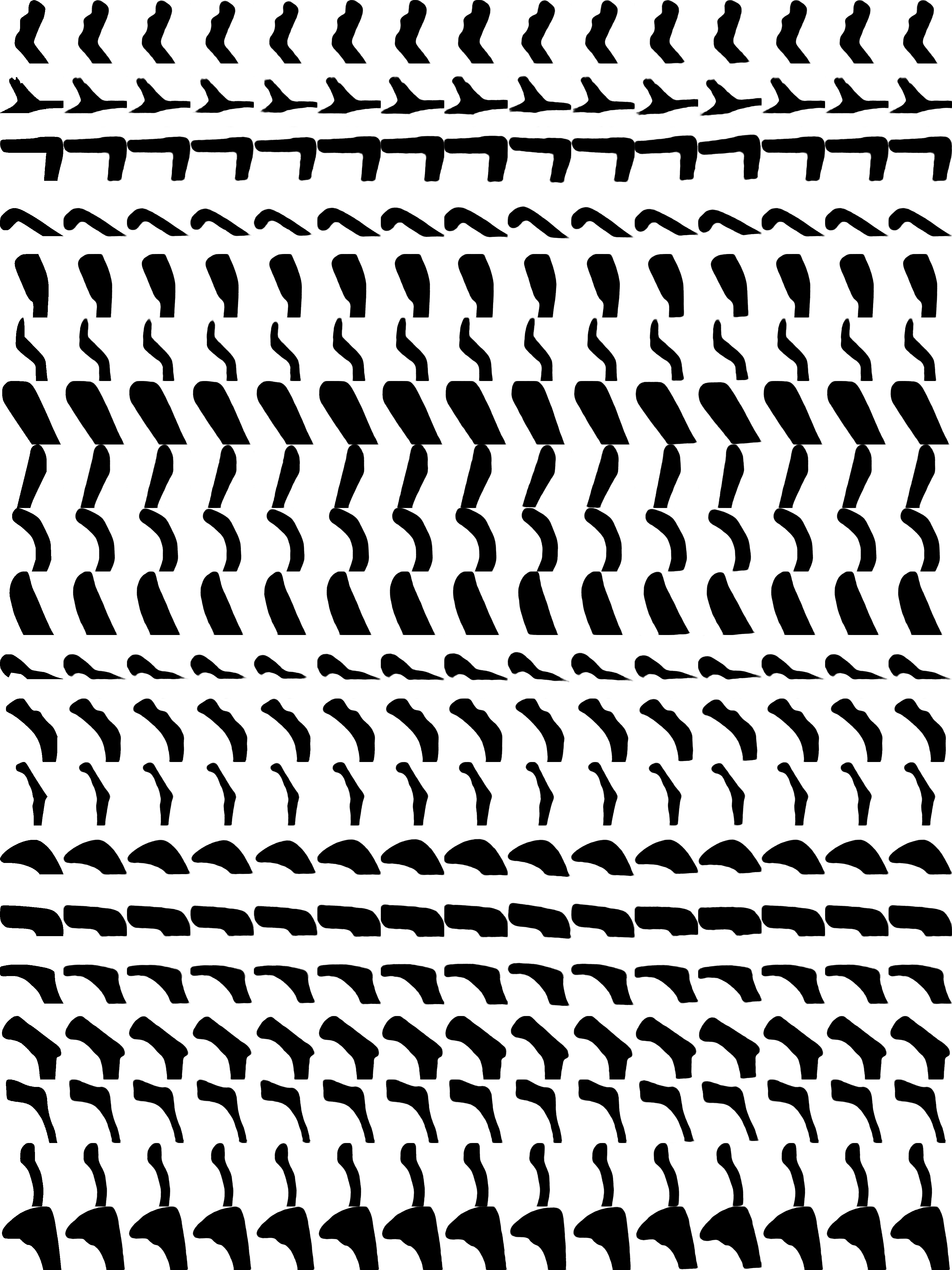}
    \hspace{1em}
    \includegraphics[width=0.545\textwidth]{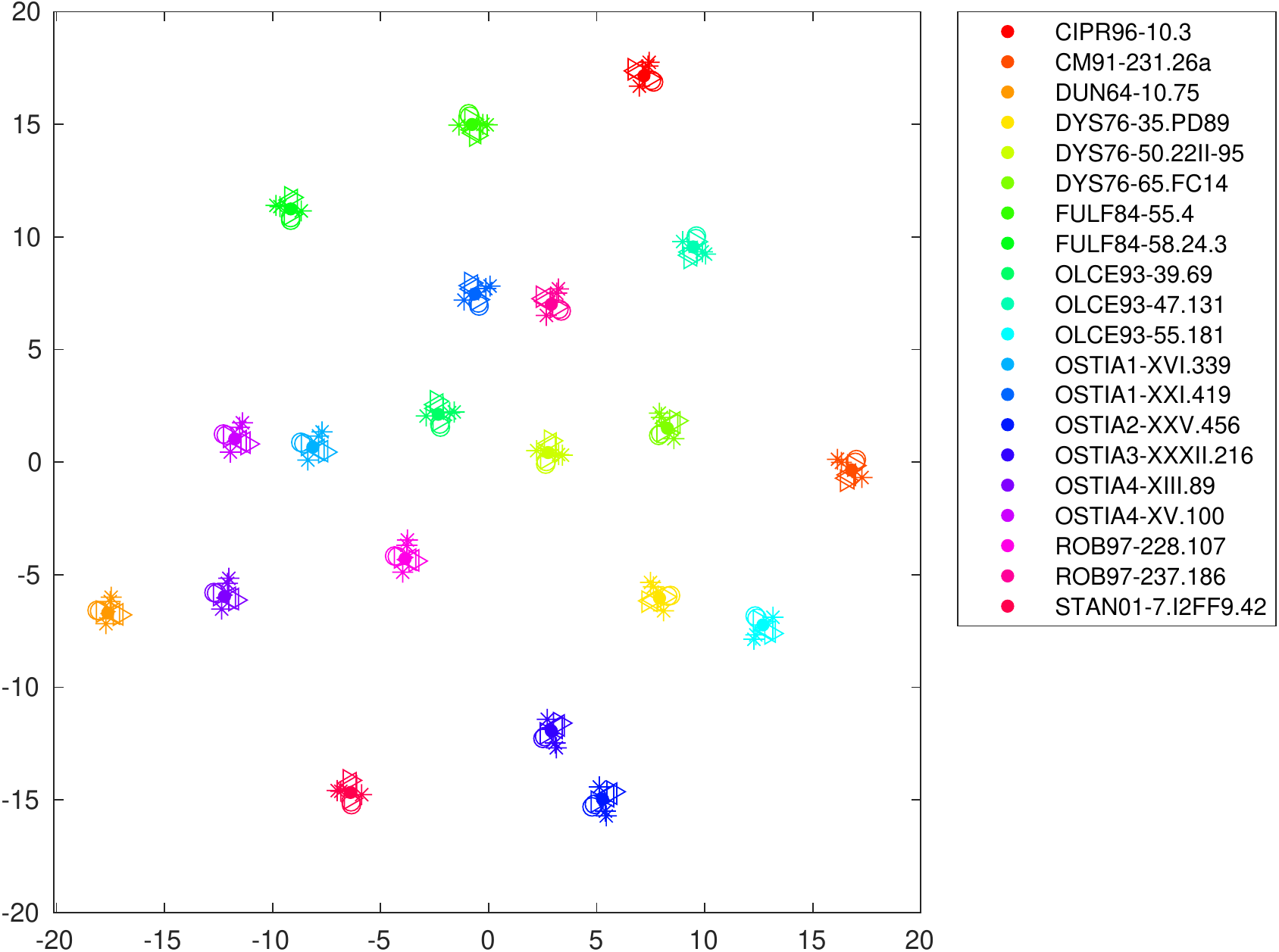}
    \caption{Robustness of latent features. Left (in columns): input shape, its reconstruction and its reconstructions \mod{after} perturbations (erosion $[3,5,7]$ pixels, dilation $[3,5,7]$ pixels and rotation $[-5,-3,3,5]$ degrees in the image space and random perturbations $[1\%,5\%,10\%]$ in the latent space); right: T-SNE plot of associated VAE features.}
    \label{fig: robustness}
\end{figure*}

Our matching assumption is based on the fact that similar shapes share similar VAE latent space representation. Thus, it \mod{is} worth to investigate if the shapes and the features in the latent space are robust to a series of \mod{perturbations} that may occur with handcrafted data.
In Figure \ref{fig: robustness} we report the reconstruction of a subset of potsherds in the ROCOPOT database under a range of \mod{perturbations} on the input shape (erosion, dilation and rotation) as well as small \mod{perturbations with a random numbers} of the parameters in latent space. In particular, we compare the reconstruction of shapes from latent space features \mod{after} undergoing {pixel erosion (of $3,5,7$ pixels), dilation (of $3,5,7$ pixels) and shape rotation (of $-5,-3,3,5$ degrees), as well as \mod{the application of a} random Gaussian perturbation \mod{on the} features in latent space ($1\%$, $5\%$ and $10\%$ \mod{of the feature values}, respectively).}
\mod{The} proximity of the latent space representations of \mod{all} shapes \mod{to their perturbed variants has been computed and visualised} in the T-SNE plot of Figure \ref{fig: robustness}, \mod{confirming the robustness of the learned features to perturbations, and the ability to match together very similar profiles. This is a visual demonstration of the feasibility of our clustering approach for the potsherds}.

\subsection{Hierarchical clustering.}
\begin{figure*}[!htb]
    \centering
    \begin{subfigure}[t]{0.1\textwidth}\centering
    \includegraphics[width=0.985\textwidth]{{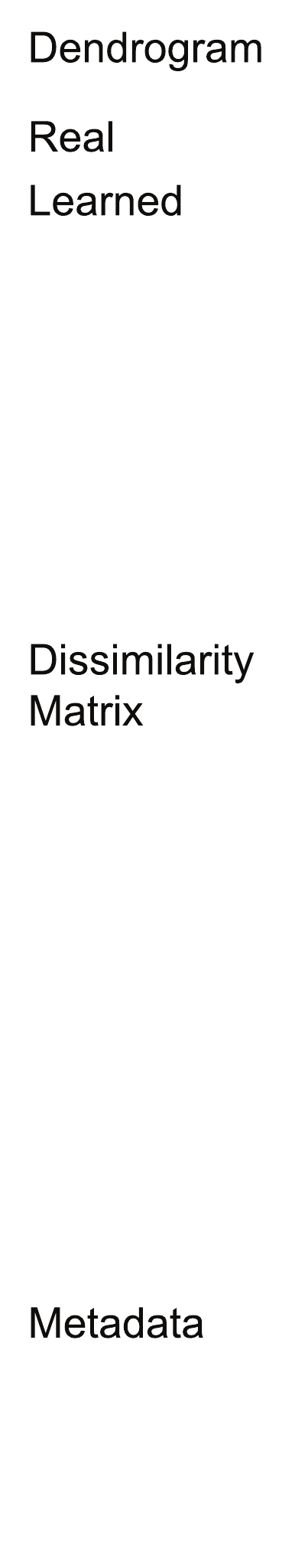}}
    \end{subfigure}
    \begin{subfigure}[t]{0.4\textwidth}\centering
    \includegraphics[width=0.985\textwidth]{{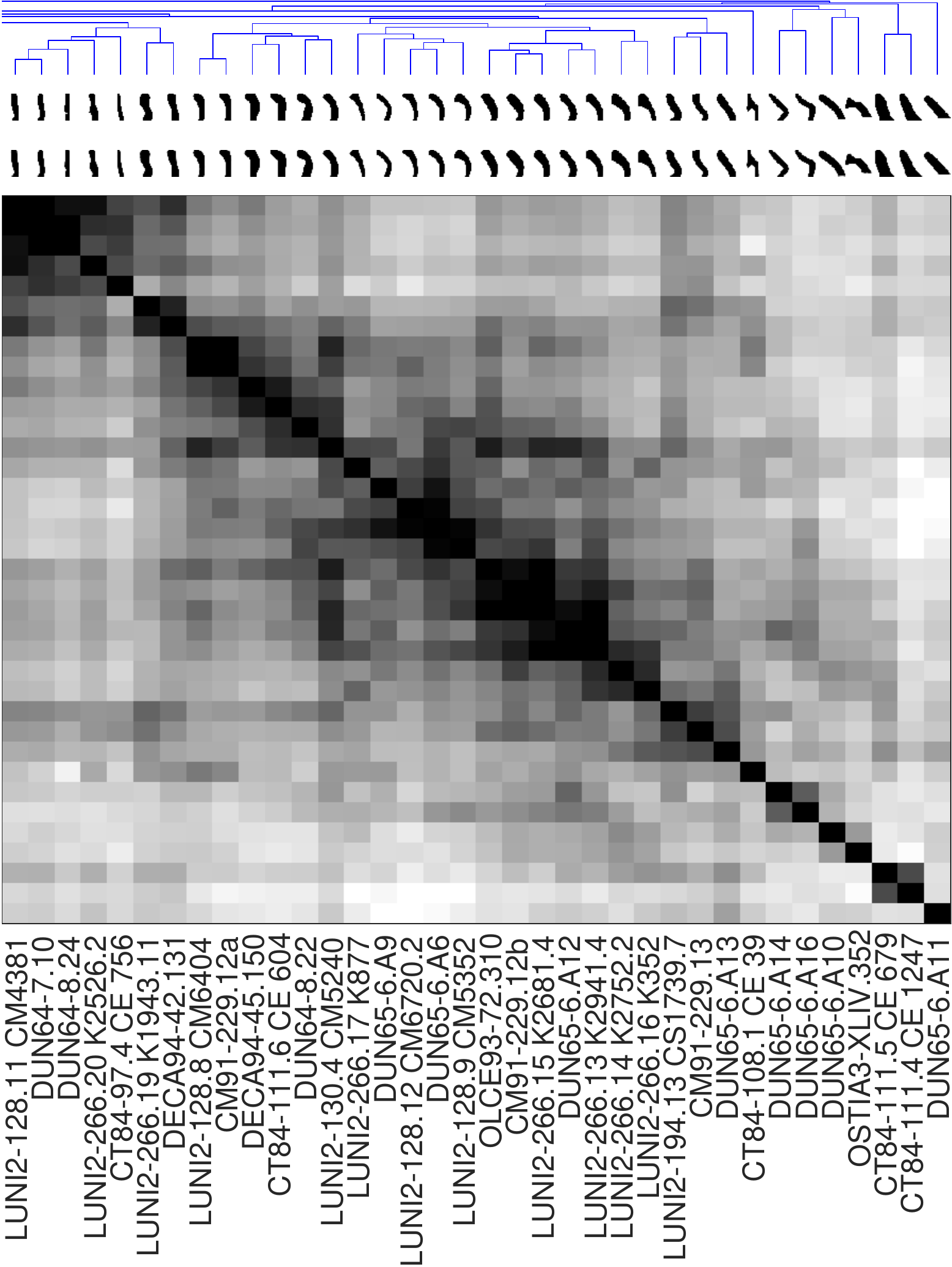}}
    \caption{Type \texttt{beaker}.}
    \label{fig: dendro-zoom BEAKER}
    \end{subfigure}
    \begin{subfigure}[t]{0.4\textwidth}\centering
    \includegraphics[width=1\textwidth]{{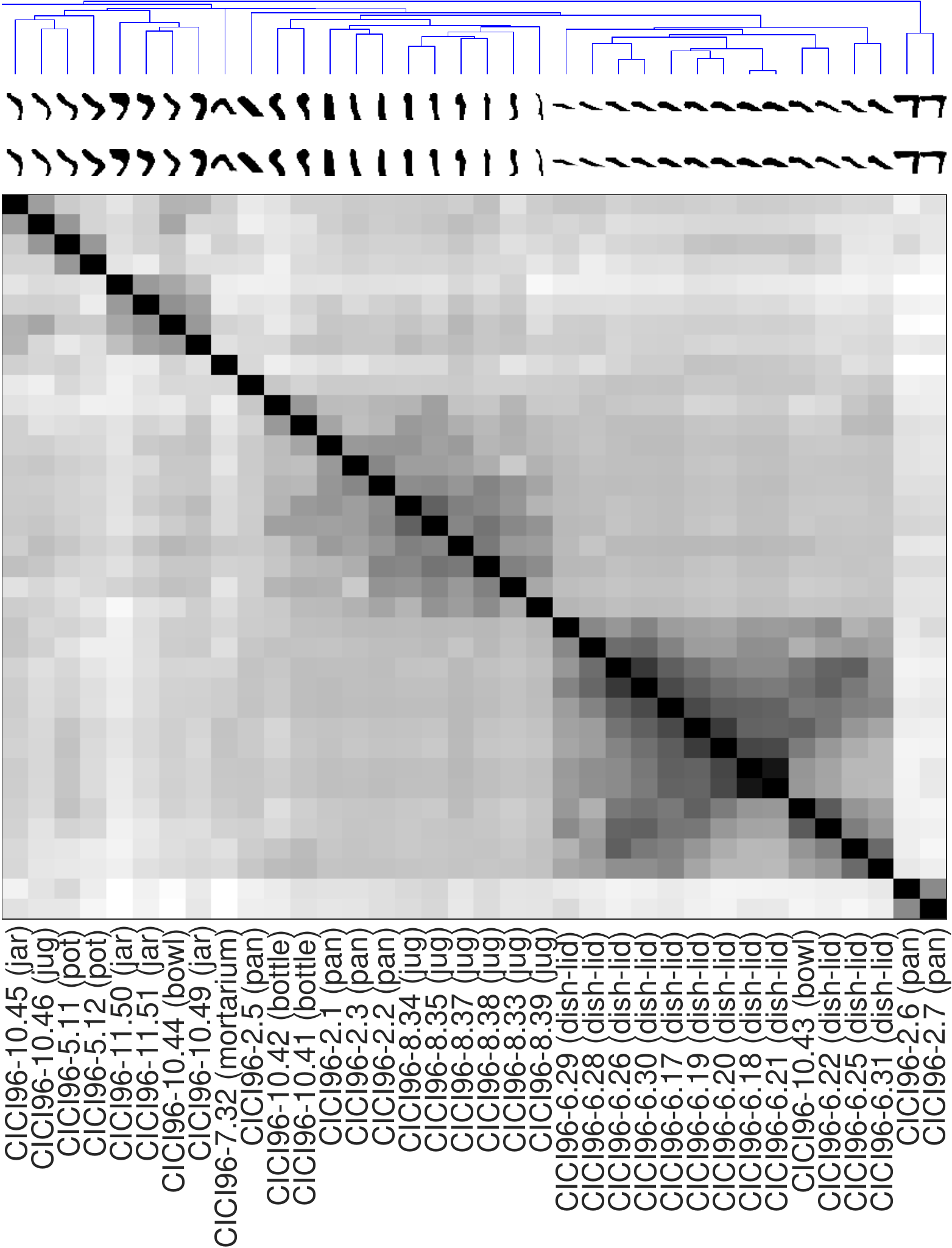}}
    \caption{Catalogue CICI96.}
    \label{fig: dendro-zoom CICI96}
    \end{subfigure}
    \caption{\mod{From top to bottom: dendrogram (partial) for type \texttt{beaker} (left) and catalogue CICI96 (right), real and learned shreds visually confirming the ability to draw shreds correctly from learned features, dissimilarity matrix and metadata.}}
\end{figure*}

As said, our approach takes advantage of the learned features in the VAE's latent space for the hierarchical clustering task.
\tocheck{In our experiments the highest cophenetic score is obtained with the choice of the \emph{Unweighted Pair Group Method using arithmetic Averages} (UPGMA, \texttt{average}) and the \mod{\texttt{Euclidean}} metric for measuring the distance between clusters.}
\tocheck{We now restrict our focus on relevant types extracted from the metadata of our dataset. These are: \texttt{domestic amphora}, \texttt{basin}, \texttt{beaker}, \texttt{bottle}, \texttt{bowl}, \texttt{censer}, \texttt{dish}, \texttt{dish-lid}, \texttt{dolium}, \texttt{jar}, \texttt{jug}, \texttt{knob}, \texttt{lid}, \texttt{mortarium}, \texttt{pan}, \texttt{pot} and \texttt{unguentarium}. For all the potsherds profiles in each type, we extracted the relevant features with the VAE strategy described before, which are hierarchically clustered in view of extracting new similarity patters. As an example, we report a portion of the hierarchical dendrogram for the type \texttt{beaker} in Figure \ref{fig: dendro-zoom BEAKER}. For convenience, we call \texttt{seed} each of these pairs, corresponding to the paired leaves at the deepest level of the dendrogram. The same dendrogram can be produced also for each single catalogue if needed to inspect intra-catalogue similarities, see Figure \ref{fig: dendro-zoom CICI96} for CICI96 \cite{CICI96}.}

\subsection{Unveiling hidden relations: the ROCOPOT App.}
\begin{figure}[!htb]
\centering
\includegraphics[width=0.475\textwidth]{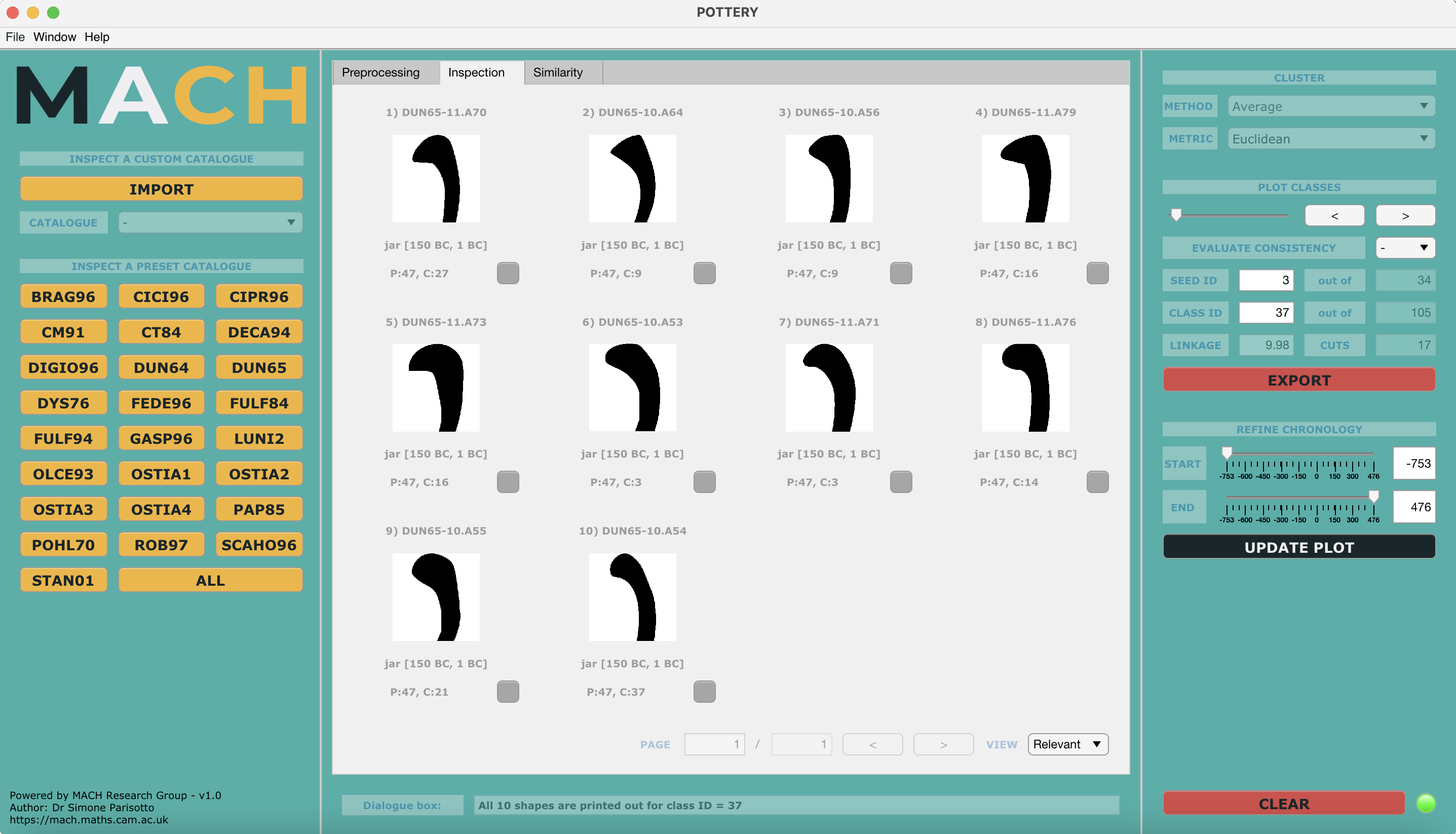}
\caption{ROCOPOT software (GUI in MATLAB R2020b).}
\label{fig: ROCOPOT app}
\end{figure}

Given the \mod{cross-disciplinary} nature of the challenge in this paper, we need a tool
for visualising and inspecting the clusters in the dendrograms. To this \mod{end}, we developed a MATLAB GUI\footnote{Freely available at \url{doi.org/10.5281/zenodo.5552265} and \url{mach.maths.cam.ac.uk/software} \cite{ROCOPOTapp}} which is able to effectively support the onfield work of the archaeologists \cite{bridges2020}, see the screenshot in Figure \ref{fig: ROCOPOT app}.

\subsection{Comments from the archaeological perspective}
In order to test how effective our workflow is in recognising actual similarities between different profiles, we inspect the quality of the dendrograms via the MATLAB app. 

The lack of a ground truth in matching potsherds, itself resulting from a degree of inherent subjectivity in the manual classification process, has a great impact on an automatic evaluation of the performances. Therefore the task of validating the matching accuracy of the dendrograms is assigned to archaeologists. Although, in principle, one could evaluate the coherence of each hierarchical group (of profiles) within the dendrogram, this task would involve a rather complex review of the results. Therefore, we asked archaeologists to concentrate on and validate the matching accuracy for each individual seed, expected to be the best match proposed by our unsupervised workflow.  

Before proceeding, we acknowledge that a large number of profiles in our dataset may in fact represent somewhat unique vessels bearing little or no similarity to any other. Roman commonwares consisted of ordinary pottery which was produced at a myriad of distinct workshops, the latter often coming into and going out of existence in the lapse of a generation. We should therefore expect the range of vessels produced, distributed and consumed, even within the same region and in the same period, to be considerably vast and varied. This otherwise unbridled drive towards a variety of forms was certainly directed - and to a degree tempered - by the practicalities of the production process (e.g.\ standardization) and the predominantly functional character of these objects. Of course, given the high levels of inter-connectivity and exchange achieved in the Roman world (and Central Tyrrhenian Italy was certainly no exception!), local potters worked within productive traditions which were used to absorb, copy and rework types originally imported from elsewhere, and whose features may have appeared from time to time better suited to specific functions or even just the tastes of potential customers. Nevertheless, only a fraction of the whole dataset may have in fact featured any actual degree of close similarity.

Our workflow processes 3678 rim profiles as featured in 25 catalogues (R in Table \ref{tab: database 1.0}) and identifies 1231 seeds (corresponding to 2462 profiles). The archaeologists looked at each seed by inspecting and comparing the paired profiles, thus assessing their similarity. They did not consider the overall shape of the vessel (even when known), but instead only concentrated on the overall combination of morphological features as presented by each individual rim (e.g.\ inclination, curve, lip design). 
This approach did not present any particular challenge to them as it was very much in keeping with standard practice in the analysis of (incomplete) potsherds as most commonly recovered through fieldwork. 
In light of the above archaeological considerations about the nature of the data itself, it is not surprising that only 500 out of 1231 seeds (41\%) were eventually validated as a positive match.
What is remarkable, however, is that out of those 500 positive seeds, only 147 (29\%) had already been indicated as close matches in the relevant pottery corpora, whilst the remaining 353 (71\%) are new. In other words, although some of these similarities have been in fact recognised and recorded when the pottery corpora were originally put together by their authors, our workflow is able to significantly expand this range of relationships. 
This goes far beyond a random matching of profiles, since the actual probability of pairing two similar profiles within the 2462 available is less than $0.001\%$.

It is certainly notable that our workflow was autonomously capable of establishing close similarity between profiles which were originally recognised as 'similar' by the original authors. This suggests that the workflow may have learned to look for the same combination of features which a pottery specialist normally looks for when classifying pottery. Although replicating such results may in itself be considered a notable achievement from a mathematical point of view, it is the fact that the range of similarities was expanded that matters a lot from an archaeological point of view. If one considers that each pottery catalogue is a reflection of relevant patterns of production, distribution and consumption within a given \mod{location} in a given period, an improved understanding of their inter-relationship can truly contribute not only to refining chronologies (useful as this exercise may be), but also to mapping how (trade) contacts between different places were created and evolved over the Roman period. Even though such patterns are routinely reconstructed on the base of the distribution of finewares and amphorae, we already pointed out that these classes may not provide as comprehensive a view of the broader population as commonwares can.

In short, the proposed workflow is very promising, especially when considering its inherent scalability. As the dataset is expanded with the inclusion of new profiles from additional pottery corpora, the learning step will improve, thus increasing the chance of recognising more (and possibly \emph{qualitatively} better) similarities. More than anything, the clustering step is capable of structuring in \emph{minutes} the kind of dataset whose size would require \emph{months} (or indeed \emph{years}!) of manual reviewing a long list of disparate publications. In fact, although here applied to commonware pottery from Central Tyrrhenian Italy, this workflow can be straightforwardly extended not only to Roman commonwares in general, but to any coherent pottery class from any time and/or period. Although this process is neither infallible or exhaustive, it is in fact meant not to replace, but rather to aid in the manual interpretation and classification of pottery profiles by unveiling \emph{additional} relationships which the mass of data may render almost \emph{invisible} to the human eye.

\section{Conclusions and Future Works} 
In this paper, we introduced the \emph{ROman COmmonware POTtery} profiles (ROCOPOT) database and provided a workflow for unveiling close similarities between (Roman commonware) pottery profiles. That is based on the hierarchical clustering of Roman potsherds via latent features learned in a deep convolutional VAE network. Such features are proved to be robust to a number of perturbations mimicking the real scenario of the high shape variance due to handcrafted materials. 
The most obvious advantage provided by our approach rests in the ability to ease and speed up the processing and matching of a very large number of pottery profiles from different corpora within a coherent and unified analytical environment. This is achieved by presenting the pottery specialist with a selection of most likely matches in the companion software, thus providing invaluable pointers not only in the development  of a comprehensive typology of Roman commonwares, but even in the future classification of potsherds coming from new excavations.
These results are very encouraging and prompt us to pursue this research further. Firstly, we plan to increase the number of profiles available in our database; secondly we will develop a specific online tool which will aid in the classification of profiles of new commonware potsherds recovered by new archaeological projects (and thus increasing our knowledge on this pottery class whilst expanding its application in the field). Finally, we aim to train the network to include other parts of the vessels (bases, main body) in its similarity search and fully incorporate the feedback (validation) of the archaeologists in order to refine its analysis through iteration.


\paragraph{Acknowledgements} 
The authors thank Michael Bronstein, Thomas Buddenkotte and Martin Millett for the useful discussions and the anonymous reviewers for their encouraging and constructive feedback. The authors acknowledge the support from the Leverhulme Trust Research Project Grant (RPG-2018-121) ``Unveiling the Invisible - Mathematics for Conservation in Arts and Humanities''. CBS further acknowledges support from the RISE projects CHiPS and NoMADS, the Cantab Capital Institute for the Mathematics of Information and the Alan Turing Institute. 

\section{Appendix}\label{sec: appendix}
Here we discuss the main concepts \mod{of} variational autoencoders and the derivation of the loss function in Equation \eqref{eq: main loss}, by means of \eqref{eq: DKL rearranged} and \eqref{eq: reconstructionloss}, see \cite{KinWel2014,doersch2021tutorial}.

We assume that the dataset consists of $\mathrm{X}=(\xbold_1,\dots,\xbold_n)$ samples, with $n>0$ (in our case $n$ two-dimensional black-white images representing the profile of potsherds of size $H\times W$ from the ROCOPOT database). 
We identify with $\xbold\in\mathrm{X}$ one sample from the dataset and with $\overline{\xbold}\in\mathrm{Y}$ the reconstruction of $\xbold$, where $\mathrm{Y}$ is the reconstruction of the space $\mathrm{X}$. Also, let $g_\phi(\blank)$ and $f_\theta(\blank)$ be the encoding and decoding functions, parametrised by $\phi$ and $\theta$, respectively. 
\mod{Moreover}, $q_\phi(\zbold\,|\,\xbold)$ represents the estimated posterior probability \mod{distribution (probabilistic encoder) for the latent code $\zbold\in\RR^k$ given $\xbold$} and $p_\theta(\xbold\,|\,\zbold)$ the likelihood of generating true data samples \mod{$\xbold$} given the latent code $\zbold$ \mod{(probabilistic decoder).  Here} $k>0$ is the fixed dimension of the reduced latent space.
\mod{In general, The Kullback-Leibler (KL) divergence $D_{\text{KL}}(q\,\|\,p)$ is a non-symmetric relative entropy designed to quantify the discrepancy between two distributions $q$ and $p$ and how much information is lost if the distribution $q$ is used to represent $p$}. 

\subsection{The Evidence Lower Bound Loss in VAE}
In the AE scheme, the idea is to learn \mod{a} low-dimensional representation $\zbold=g_\phi(\xbold)\in\RR^k$ of the data by means of learning the identity function such that $\overline{\xbold}=f_\theta(g_\phi(\xbold))$. The overfitting and the impossibility to generate new samples are clear limits of this approach.

Conversely, VAEs aim to learn the model underlying the data: the estimated posterior $q_\phi(\zbold\,|\,\xbold)$ \mod{is designed to be close} to the real \mod{posterior}  $p_\theta(\zbold\,|\,\xbold)$ making possible to generate true samples. \mod{To that effect, the \emph{forward} or \emph{reverse} KL divergence could be used.} The \emph{forward} KL divergence $D_\text{KL}(p_\theta(\zbold\,|\,\xbold)\,\|\,q_\phi(\zbold\,|\,\xbold))$ tends to \mod{favour an approximate} distribution $q_\phi$ with an undesired mean effect; conversely the \emph{reversed} KL divergence $D_\text{KL}(q_\phi(\zbold\,|\,\xbold)\,\|\,p_\theta(\zbold\,|\,\xbold))$ is better suited to approximate the distribution 
$q_\phi$ within a mode of $p_\theta$ and is preferred \cite{TheOorBet2016,huszar2015not,goodfellow2017nips,nguyen2017dual}. 
\mod{The} \emph{reversed} KL divergence is equivalent to:
\[
\begin{aligned} \allowdisplaybreaks
D_\text{KL}&(q_\phi(\zbold\,|\,\xbold)\,\|\, p_\theta(\zbold\,|\,\xbold) ) = \int q_\phi(\zbold \,|\, \xbold) \log\frac{q_\phi(\zbold \,|\, \xbold)}{p_\theta(\zbold \,|\, \xbold)} \diff\zbold & \\ 
=&\int q_\phi(\zbold \,|\, \xbold) \log\frac{q_\phi(\zbold \,|\, \xbold)p_\theta(\xbold)}{p_\theta(\zbold, \xbold)} \diff\zbold\\ 
=&\int q_\phi(\zbold \,|\, \xbold) \left( \log p_\theta(\xbold) + \log\frac{q_\phi(\zbold \,|\, \xbold)}{p_\theta(\zbold, \xbold)} \right) \diff \zbold\\ 
=&\log p_\theta(\xbold) + \int q_\phi(\zbold \,|\, \xbold)\log\frac{q_\phi(\zbold \,|\, \xbold)}{p_\theta(\zbold, \xbold)} \diff\zbold\\ 
=&\log p_\theta(\xbold) + \int q_\phi(\zbold \,|\, \xbold)\log\frac{q_\phi(\zbold \,|\, \xbold)}{p_\theta(\xbold\,|\,\zbold)p_\theta(\zbold)} \diff\zbold \\ 
=&\log p_\theta(\xbold) + \mathbb{E}_{\zbold\sim q_\phi(\zbold \,|\, \xbold)}\left(\log \frac{q_\phi(\zbold \,|\, \xbold)}{p_\theta(\zbold)} - \log p_\theta(\xbold \,|\, \zbold)\right)\\ 
=&\log p_\theta(\xbold) + D_\text{KL}(q_\phi(\zbold\,|\,\xbold) \,\|\, p_\theta(\zbold)) \\&- \mathbb{E}_{\zbold\sim q_\phi(\zbold\,|\,\xbold)}\log p_\theta(\xbold\,|\,\zbold),
\end{aligned}
\]
where in the above we used $p_\theta(\zbold \,|\,\xbold) = p_\theta(\zbold,\xbold) / p_\theta(\xbold)$, $\int q_\phi(\zbold \,|\,\xbold) \diff\zbold = 1$ and $p_\theta(\zbold,\xbold) = p_\theta(\xbold\,|\,\zbold) p_\theta(\zbold)$. 
By rearranging the above equation, we have:
\begin{equation}
\begin{aligned}
&\log p_\theta(\xbold) - D_\text{KL}( q_\phi(\zbold\,|\,\xbold) \,\|\, p_\theta(\zbold\,|\,\xbold) ) 
\\
&= \mathbb{E}_{\zbold\sim q_\phi(\zbold\,|\,\xbold)}\log p_\theta(\xbold\,|\,\zbold) - D_\text{KL}(q_\phi(\zbold\,|\,\xbold)\,\|\,p_\theta(\zbold))
\end{aligned}
\label{eq: rearranging ELBO}
\end{equation}
The left-hand-side of Equation \eqref{eq: rearranging ELBO} allows us to maximise the (log-)likelihood of generating real data, i.e.\ $\log p_\theta(\xbold)$, while minimising the difference between the real and estimated posterior distributions.
Note that $p_\theta(\xbold)$ is fixed with respect to $q_\phi(\zbold\,|\,\xbold)$.
The negative of the left-hand-side in Equation \eqref{eq: rearranging ELBO} is defined as the \emph{Evidence Lower Bound} (ELBO) loss function:
\begin{equation}
\begin{aligned}
\mathcal{L}_\text{ELBO}(\theta, \phi)
=& 
-\mathbb{E}_{\zbold\sim q_\phi(\zbold\,|\,\xbold)}\log p_\theta(\xbold\,|\,\zbold)\\
&+ 
D_\text{KL}(q_\phi(\zbold\,|\,\xbold)\,\|\,p_\theta(\zbold)).
\end{aligned}
\label{eq: ELBO loss}
\end{equation}

The optimal parameters are obtained by solving the following minimisation problem:
\[
(\theta^{*},\phi^{*}) 
= 
\arg\min_{\theta, \phi} \mathcal{L}_\text{ELBO}(\theta,\phi).
\]
Being the reverse KL divergence always non-negative, -$\mathcal{L}_{\text{ELBO}}$ is the lower bound of $\log p_\theta (\xbold)$: this explains why the minimisation of the $\mathcal{L}_{\text{ELBO}}$ loss implies the maximisation of the probability of generating real data samples $p_\theta(\xbold)$. Indeed, from Equation \eqref{eq: rearranging ELBO} we have:
\[
\begin{aligned}
-\mathcal{L}_{\text{ELBO}} 
&= \log p_\theta(\xbold) - D_\text{KL}( q_\phi(\zbold\,|\,\xbold)\,\|\,p_\theta(\zbold\,|\,\xbold) )\\
&\leq \log p_\theta(\xbold).
\end{aligned}
\]

\subsection{Derivation of the KL divergence}\label{sec: reversed KLdiv}
The KL divergence in Equation \eqref{eq: ELBO loss} is a \emph{regulariser} because it is a constraint on the form of the approximate posterior \cite{odaibo2019tutorial}.
In general, the KL divergence of two multivariate Gaussian distributions $q_\phi$ and $p_\theta$ (of dimension $n$), i.e.\ $q_\phi = \mathcal{N}(\bm{\mu}_\phi,\bm{\sigma}_\phi)$ and $p_\theta = \mathcal{N}(\bm{\mu}_\theta,\bm{\sigma}_\theta)$, can be expressed in the following form:
\begin{equation}
\begin{aligned}
D_\text{KL}(q_\phi\mid\mid p_\theta) 
=&
\frac{1}{2}\Bigg(\log\frac{|\bm{\sigma}_2|}{|\bm{\sigma}_1|} - n
+ \text{tr}\left( \bm{\sigma}_2^{-1}\bm{\sigma}_1 \right) \\&+ (\bm{\mu}_2 - \bm{\mu}_1)^T \bm{\sigma}_2^{-1}(\bm{\mu}_2 - \bm{\mu}_1)\Bigg).
\label{eq: DKL def}
\end{aligned}
\end{equation}

In VAE, $q_\phi$ is the encoder distribution defined as $q_\phi(\zbold\,|\,\xbold)=\mathcal{N}(\zbold\,|\,\bm{\mu}(\xbold),\bm{\sigma}(\xbold))$ where $\bm{\sigma}=\text{diag}(\sigma_1^2,\dots,\sigma^2_n)$ while $p_\theta$ is the latent prior given by $p_\theta(\zbold)=\mathcal{N}(\bm{0},\Ibold)$. 
Thus, from Equation \eqref{eq: DKL def} the reverse KL divergence $D_\text{KL}(q_\phi(\zbold\,|\,\xbold)\,\|\, p_\theta(\zbold))$ is expressed as:
\[
\begin{aligned}
&=\frac{1}{2}\left(\log\frac{|\Ibold|}{|\bm{\sigma}|} - n + \text{tr} \{\Ibold^{-1}\bm{\sigma} \} + (\bm{0} - \bm{\mu})^T \Ibold^{-1}(\bm{0} - \bm{\mu})\right)\\
&=\frac{1}{2}\left(-\log{|\bm{\sigma}|} - n + \text{tr} \{ \bm{\sigma} \} + \bm{\mu}^T \bm{\mu}\right)\\
&=\frac{1}{2}\left(-\log\prod_{i=1}^k\sigma_i^2 - n + \sum_{i=1}^k\sigma_i^2 + \sum_{i=1}^k\mu^2_i\right)\\
&=\frac{1}{2}\left(-\sum_{i=1}^k\log\sigma_i^2 - n + \sum_{i=1}^k\sigma_i^2 + \sum_{i=1}^k\mu^2_i\right)
\end{aligned}
\]
which can be rearranged as
\begin{equation}\small
D_\text{KL}(q_\phi(\zbold\,|\,\xbold)\,\|\, p_\theta(\zbold))=  -\frac{1}{2}\sum_{i=1}^k\left(\log\sigma_i^2 + 1 - \sigma_i^2 - \mu^2_i\right).
\label{eq: DKL rearranged}
\end{equation}

\subsection{Reconstruction loss}
The expectation in Equation \eqref{eq: ELBO loss} is a \emph{reconstruction} loss. In our scenario, similar to a semantic segmentation problem with labels $0$ and $1$ due to the black-white nature of the data, it makes sense to use a sigmoid cross-entropy with logits energy term as reconstruction loss $\mathcal{L}_c(\xbold,\overline{\xbold})$, of the form
\[
- \xbold \cdot \log(\sigmoid(\overline{\xbold})) - (\bm{1}-\xbold)  \cdot\log(\bm{1} - \sigmoid(\overline{\xbold})),
\]
where $\sigmoid$ is the sigmoid function, reformulated for numerical stability and for avoiding overflows\footnote{\url{www.tensorflow.org/api_docs/python/tf/nn/sigmoid_cross_entropy_with_logits}} as
\begin{equation}
\mathcal{L}_c(\xbold,\overline{\xbold}) = \max(\overline{\xbold}, \bm{0}) - \overline{\xbold} \cdot \zbold + \log\left(\bm{1} + \exp\left(-|\overline{\xbold}|\right)\right).
\label{eq: reconstructionloss}
\end{equation}

\subsection{Reparametrisation trick}
\label{sec: reparametrisation}
The expectation term in Equation \eqref{eq: ELBO loss} requires generating samples from the latent space $q_\phi(\zbold\,|\,\xbold)$. However, the sampling is a stochastic process, making impossible to backpropagating the gradient. 
The reparameterisation trick proposed in \cite{KinWel2014} (based on methods already introduced in \cite{devroye1986,Devroye96})
allows to overcome such difficulties by means of rewriting $\zbold$ as a sum of a deterministic part plus an independent random variable $\bm{\varepsilon}\sim\mathcal{N}(\bm{0},\bm{\Ibold})$:
\begin{equation}
\zbold=\bm{\mu}+\bm{\sigma}\odot\bm{\varepsilon},
\label{eq: reparametrisation}
\end{equation}
where $\odot$ refers to element-wise product.
With this trick in place, the mean $\bm{\mu}$ and the variance $\bm{\sigma}$ of the distribution are now learnable parameters.

\bibliographystyle{elsarticle-num} 
\bibliography{biblio}
\end{document}